\documentclass[10pt,twocolumn,letterpaper]{article}

\usepackage[pagenumbers]{cvpr} % To force page numbers, e.g. for an arXiv version

\usepackage[accsupp]{axessibility}  % Improves PDF readability for those with disabilities.

\usepackage[linesnumbered,ruled,vlined]{algorithm2e}

\DeclareMathOperator*{\argmin}{arg\,min}
\definecolor{cvprblue}{rgb}{0.21,0.49,0.74}
\usepackage[pagebackref,breaklinks,colorlinks,allcolors=cvprblue]{hyperref}

\def\paperID{*****} % *** Enter the Paper ID here
\def\confName{CVPR}
\def\confYear{2026}

\title{Instant Colorization of Gaussian Splats}
\author{Daniel Lieber\\
University of Osnabrück\\
{\tt\small dlieber@uos.de}
\and
Alexander Mock\\
University of Osnabrück\\
{\tt\small alexander.mock@uos.de}
\and
Nils Wandel\\
University of Osnabrück\\
{\tt\small nils.wandel@uos.de}
}

\begin{document}
\maketitle
\begin{abstract}
Gaussian Splatting has recently become one of the most popular frameworks for photorealistic 3D scene reconstruction and rendering. While current rasterizers allow for efficient mappings of 3D Gaussian splats onto 2D camera views, this work focuses on mapping 2D image information (e.g. color, neural features or segmentation masks) efficiently back onto an existing scene of Gaussian splats.
This 'opposite' direction enables applications ranging from scene relighting and stylization to 3D semantic segmentation, but also introduces challenges, such as view-dependent colorization and occlusion handling.

Our approach tackles these challenges using the normal equation to solve a visibility-weighted least squares problem for every Gaussian and can be implemented efficiently with existing differentiable rasterizers. We demonstrate the effectiveness of our approach on scene relighting, feature enrichment and 3D semantic segmentation tasks, achieving up to an order of magnitude speedup compared to gradient descent-based baselines.
\end{abstract}
\section{Introduction}
\label{sec:intro}

Recently, scene representations based on Gaussian splats (GS) have achieved notable success in photorealistic novel view synthesis due to their high computational efficiency and their ability to render complex scenes in real time with high detail \cite{kerbl3Dgaussians,Huang2DGS2024,dontsplat}.
By optimizing the Gaussian parameters via gradient descent, the scene appearance can be iteratively refined to closely match a set of input camera views. 
However, after a scene has been reconstructed, modifying its appearance or enriching the representation with semantic features remains a relevant but also challenging task as it must account for occlusions caused by partially transparent splats as well as view-dependent effects. 
Although gradient-based optimization can solve this problem, it typically requires many iterations and substantial computational cost.

\begin{figure*}[h!]
    \centering
    \includegraphics[width=0.9\linewidth]{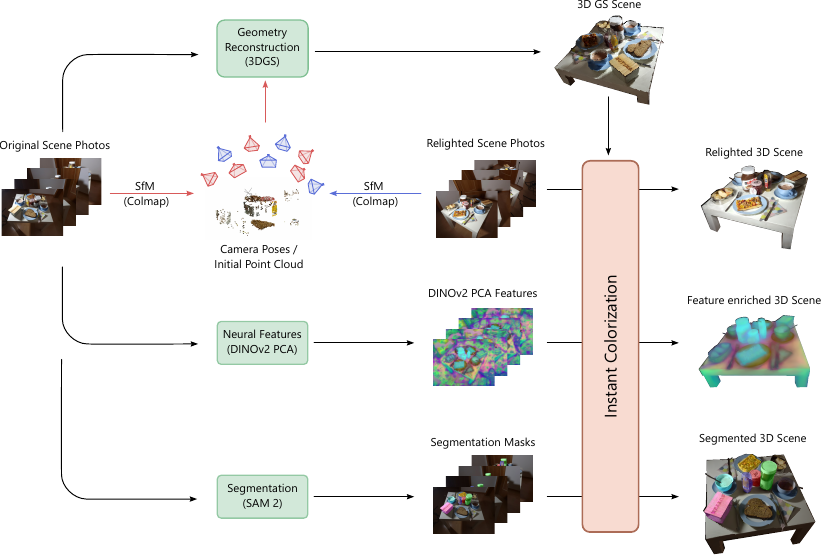}
    \caption{Architecture overview: First, camera poses and an initial point cloud are generated from a set of original scene photos using structure from motion (COLMAP) \cite{schoenberger2016mvs,schoenberger2016sfm}. Then, the 3DGS framework \cite{kerbl3Dgaussians} is used to generate a 3D Gaussian scene representation. This work focuses on mapping new 2D image information onto this existing GS scene (see red box). Therefore, three scenarios are considered: 
    First, we capture the same scene a second time under different lighting conditions in order to relight the Gaussian splats. To ensure correct camera pose registration with the 3D scene, COLMAP is used (see blue arrow). 
    Second, we extract principal components from DINOv2 features \cite{oquab2023dinov2} of the original photos to obtain a feature enriched 3D scene representation. 
    Finally, we generate segmentation masks using the SAM2 model \cite{ravi2025sam2} to do 3D semantic scene segmentation (bottom row).}
    \label{fig:architecture_overview}
\end{figure*}

%Thus, we aim to update the appearance parameters of an existing scene representation by efficiently mapping a new set of images, e.g., captured under different lighting conditions, containing neural features or segmentation masks onto the Gaussian splats. 

Thus, we propose a novel algorithm to efficiently map 2D images onto existing 3D Gaussian splat scenes and apply it in 3 different settings (see \Cref{fig:architecture_overview}): First, scene relighting, second, feature enrichment of scene representations and third, 3D scene segmentation. 
The algorithm leverages the normal equation to solve a visibility-weighted least squares problem between rendered and target images. It explicitly accounts for occlusions, alpha blending of partially transparent splats, and view-dependent colors, and can be easily implemented based on existing differentiable GS rendering frameworks. %Beyond relighting via light baking, texture projection is also relevant for applications such as texture mapping, scene stylization, feature enrichment, and semantic scene segmentation.

% Our main contributions are:
% \begin{enumerate}
% \item A visibility-aware least squares formulation for projecting 2D image information onto existing 3D Gaussian splat scenes that explicitly accounts for visibility and view-dependent appearance. % Novel Method
% \item An efficient solution using the normal equation that enables rapid colorization of existing Gaussian splat representations without additional training. % Novel Speed
% %\item A general projection mechanism that allows integrating diverse image-derived signals, including textures, neural feature maps, and semantic segmentation masks, into Gaussian splat scenes. % And you can also apply it to downstream tasks
% %\textcolor{red}{evtl relighting dataset erwähnen?}
% %\item Evaluations on three different tasks, including scene relighting, feature enrichment of scene representations and 3D scene segmentation.
% \item Multiple demonstrations that show the general applicability of \emph{instant colorization} on three downstream tasks, including scene relighting, feature enrichment of scene representations and 3D scene segmentation.
% \end{enumerate}

%We evaluate the proposed approach on scene relighting, 3D semantic scene segmentation, and neural feature enrichment tasks.
To this end, we recorded additional scenes under varying lighting conditions and generated neural features as well as semantic segmentation masks using the DINOv2 \cite{oquab2023dinov2} and the Segment Anything Model 2 \cite{ravi2025sam2} respectively.
The proposed colorization algorithm achieves up to an order of magnitude faster 2D to 3D projections compared to gradient descent-based optimization approaches like e.g. Adam \cite{kingma2017adammethodstochasticoptimization}. Code and data are available on github \url{https://github.com/dlieber01/Instant-Colorization-of-Gaussian-Splats}.

In the following, we first give a short overview over related work. Then, we derive the colorization algorithm and explain our experimental setup. Finally, we present our results and conclude with a short discussion. 
\section{Related Work}
\label{sec:related_work}
Generating highly realistic 3D scene representations from photographs and rendering novel views efficiently is a long standing research problem in computer vision~\cite{liu2025survey}.
With the upcoming of gradient based optimization schemes for scene representations based on differentiable rasterizers~\cite{liu2019softras,laine2020modularprimitiveshighperformancedifferentiable,Mitsuba3} or neural radiance fields~\cite{mildenhall2020nerf, mueller2022instant}, realism has been significantly improved.
While volumetric representations based on Gaussians have been studied already for several decades~\cite{Westover1991SplattingAP,ewa_volume_splatting,hw_ewa_splatting,rhodin2015versatile}, Kerbl et al. \cite{kerbl3Dgaussians} have recently shown that representations based on Gaussian splats enable highly efficient novel view synthesis on high-fidelity scenes that can be optimized via gradient-descent. 
Optimization improvements via more efficient rasterizers~\cite{ye2025gsplat}, the Levenberg-Marquardt algorithm (3DGS-LM~\cite{hoellein_2025_3dgslm}) or regularization-based training~\cite{chen2025quantifying} have further reduced training time and improved robustness. 
The success of Gaussian Splatting has already inspired countless follow-up works \cite{Dalal_2024}, e.g. methods that make use of 2D Gaussian splats~\cite{Huang2DGS2024}, more accurate perspective projections \cite{hahlbohm2025htgs}, or ray-traced variants for physically-based rendering~\cite{moenneloccoz2024,dontsplat}.
The popularity of gaussian splats for scene representations motivates the search for efficient texture projection techniques that allow to map 2D image information such as relighted scene photos, neural features, segmentation masks, etc back onto the corresponding splats in 3D.

\paragraph{Relightable Gaussian Splatting}
Several works focus on relightable Gaussian splatting that incorporate surface normals, BRDF parameters and illumination maps (e.g., Relightable 3D Gaussians \cite{Gao2024Relightable3DGaussian} or LumiGauss \cite{Kaleta2025LumiGauss}).
Often, these representations not only allow to change the illumination but also model reflections~\cite{Gao2024Relightable3DGaussian, NormalGS2024}.
However, more complicated light interactions such as caustics or subsurface scattering are still out of reach. 
%In contrast, classical texture-projection approaches based on projective mapping or ray casting provide fast appearance transfer but typically ignore view-dependent appearance and detailed visibility reasoning. % den satz verstehe ich nicht
Here, data-driven approaches that capture a scene under different light conditions can become a viable option \cite{debevec1998lightstage}.
While deep-learning based approaches achieve impressive results for example for subsurface scattering \cite{Dihlmann2024SSSGS} or human avatars \cite{Sun2025GRGS}, 
training neural networks is computationally expensive and might be prone to overfitting if the dataset is not large enough. 
Thus, in case of scarce data and computational budget, an efficient image mapping strategy might be more suitable \cite{sun_light_stage_superres_2020}.

\paragraph{Textured Gaussian Splatting}
Textured-GS \cite{chao2025texturedgaussians, huang2024texturedgsgaussiansplattingspatially} and subsequent approaches have shown that extending the standard spherical-harmonic (SH) color representation with spatially varying or learned color fields improves appearance quality and editability. 
This decoupling of geometry and appearance aims to bridge the gap between neural volumetric rendering and traditional mesh-based texturing resulting in higher-quality renderings. 
Further extensions such as Spec-Gaussian~\cite{yang2024spec} and SuperGaussians~\cite{xu2024SuperGaussians} enhance the expressiveness of individual splats by modeling view-dependent reflectance or spatially varying color distributions. 
Unfortunately, non-linear color models significantly complicate direct texture projections. Therefore, this work only focuses on linear color models, such as those based on spherical harmonics (SH), which still account for a large proportion of current research into Gaussian splats \cite{ye2025gsplat,Huang2DGS2024}.

\paragraph{Feature-Enriched 3D Scene Representations}
Beyond discrete semantic labels, recent work has demonstrated that dense image features from vision or vision-language models can be integrated into spatial scene representations.
Such feature-enriched maps enable queryable scene representations that support multimodal interaction, most prominently through natural language \cite{lerf2023, peng2023openscene, maggio2024clio, garfield2024}.
Other approaches further exploit these representations for open-set mapping, scene graphs, and spatial reasoning \cite{conceptfusion2023, clipfields2023, guenther2026sgb, igelbrink2026disc}.

These methods typically require specialized neural scene representations or additional training procedures to associate image features with the underlying geometry.
In contrast, our approach directly projects dense feature maps onto an existing Gaussian splat representation, enabling feature-enriched scene geometry without additional learning and with low computational overhead.

% \textcolor{red}{TODO: Motivation feature enrichment: viele arbeiten brauchen das downstream...}
% Alex: Ich habe den 3D Scene Segmentation mit dem Textured Gaussian Splats paragraph getauscht. Das ist vielleicht etwas passender mit der Reihenfolge der Experimente
% Außerdem habe ich das Feature Enrichment motivitiert. Ist jetzt etwas doppelt mit dem Text der Experimente. Evtl sollte der Text in den Experimenten gekürzt werden
% Sonst: Ich mache am Ende das related Work immer gerne mit einem kleinen Absatz, der zusammenfasst, was im Related Work fehlt und/oder wo wir drauf aufbauen. Aber ist gerade zumindest schon da

\paragraph{3D Scene Segmentation}
has important applications in robotics \cite{Ming_2025}, autonomous driving \cite{chen2025snerfautonomousdrivingsimulation}, and semantic scene understanding \cite{engelmann2024opennerf}.
Recent advances have shown that Gaussian-based scene representations can also be extended for semantic segmentation and object editing. Semantic-GS \cite{guo2024semanticgaussiansopenvocabularyscene} leverages 2D segmentation priors from foundation models such as SAM \cite{kirillov2023segany} and SAM2 \cite{ravi2025sam2} to propagate semantic labels into 3D Gaussian fields.
RT-GS2 \cite{jurca2024rtgs2realtimegeneralizablesemantic} enables real-time semantic segmentation of Gaussian splat representations. 
However, these approaches typically require training additional neural networks to infer semantic labels from the 3D representation. 
Segment Any Gaussian (SAGA) \cite{cen2025saga} learns a contrastive feature field for Gaussian splatting representations to enable interactive segmentation queries. 
This requires an additional training stage to associate semantic features with the Gaussian splats.

In contrast, our goal is to directly leverage high-quality 2D segmentation models such as SAM2 \cite{ravi2025sam2} by efficiently projecting their outputs into the 3D Gaussian scene representation, avoiding the need for additional training. 
\section{Method}
\label{sec:method}
This section derives a novel method for colorizing Gaussian splats that can be easily implemented with existing differentiable renderers. 
To this end, we formulate the colorization problem as a weighted least squares problem, which can be solved using the normal equation. 
The final algorithm is outlined in \cref{alg:colorization}. \Cref{fig:architecture_overview} shows, how this colorization algorithm can be used for various tasks such as scene relighting, feature enrichment or 3D scene segmentation.
%TODO: etwas schöner formulieren. auch neural feature augmentation erwähnen

% Differentiable renderer
\paragraph{Differentiable renderer} 
To optimize a scene representation of Gaussian splats, typically a differentiable renderer creates images from a set of $N_g$ Gaussian splats from different camera perspectives:
\begin{equation}
    \textrm{img}_{x,y}^j = \textrm{render}(P_j,(\vec \mu_i,\Sigma_i,c_i^m)_{i=1,...,N_g}^{m=1,...,(L+1)^2})
\end{equation}
Here, $\textrm{img}_{x,y}^j$ denotes the pixels of the rendered image with screen coordinates $x,y$ from the perspective of camera $j$. $P_j$ denotes the $j\textsuperscript{th}$ camera parameters, $\mu_i$ and $\Sigma_i$ denote the means and covariances of the $i\textsuperscript{th}$ Gaussian splats and $c_i^m$ denotes its $m\textsuperscript{th}$ SH coefficients to describe view dependent colors. 
Following Kerbl et al. \cite{kerbl3Dgaussians}, we can compute the value of a pixel $\textrm{img}_{x,y}^j$ by $\alpha$-blending a depth ordered list of $\mathcal{N}$ gaussian splats that overlap the pixel:
\begin{equation}
    \textrm{img}_{x,y}^j = \sum_{i = 1 }^\mathcal{N}\underbrace{\left(\prod_{k=1}^{i-1}(1-\alpha_k) \right)}_{\textrm{occlusions}} \alpha_i C_i(\vec d)
    \label{eq:alpha_blend}
\end{equation}
Here, $\alpha_i$ is the transparency of the $i\textsuperscript{th}$ Gaussian splat projected onto pixel $(x,y)$ and $C_i(\vec d)$ is the color of the Gaussian from viewing direction $\vec d$. 
If we consider spherical harmonics up to a maximum degree of $L$, there are $(L+1)^2$ SH basis functions $Y_m(\vec d)$ and the color of a Gaussian $i$ from a viewing direction $\vec d$ can be computed as follows:

\begin{equation}
    C_i(\vec d) = \sum_{m=1}^{(L+1)^2}c_i^m Y_m(\vec d)
    \label{eq:SH_coeff}
\end{equation}

Note, that for the sake of clarity, we only consider a single channel here, however, by extending $\textrm{img}_{x,y}^{j,k}$ and $c_i^{m,k}$ with another channel index $k$, all of the following derivations can be easily extended to multiple channels (e.g. RGB color channels, neural features or segmentation masks).

% Visibility
\paragraph{Visibility} 
The visibility $V_{i,x,y}^j$ of a Gaussian $i$ within a rendered pixel $\textrm{img}_{x,y}^j$ can be computed by taking the partial derivative of $\textrm{img}_{x,y}^j$ with respect to the Gaussian's color $C_i(\vec d)$. 
As can be seen from \cref{eq:alpha_blend}, this derivative automatically incorporates visibility-affecting factors, such as transparency $\alpha_i$ and occlusions. 
By applying the chain rule and 
considering that the first spherical harmonic is constant ($Y_1(\vec d)=1/\sqrt{4\pi}$), we obtain the following expression for $V_{i,x,y}^j$:

\begin{equation}
\resizebox{\columnwidth}{!}{$
    V_{i,x,y}^j = \frac{\partial \textrm{img}_{x,y}^j}{\partial C_i(\vec d)} = \frac{\partial \textrm{img}_{x,y}^j}{\partial C_i(\vec d)} \underbrace{ \frac{\partial C_i(\vec d)}{\partial c_i^1}}_{=Y_1(\vec d)=1/\sqrt{4\pi}}\sqrt{4\pi} = \frac{\partial \textrm{img}_{x,y}^j}{\partial c_i^1}\sqrt{4\pi}
    \label{eq:pixel_visibility}
$}
\end{equation}

%\todo{besser erklären! vllt kombinieren mit Eq 8}
Now, by summing over all pixels, we can compute the visibility of a Gaussian $i$ whithin the entire camera image $j$:
\begin{equation}
    V_i^j = \sum_{x,y}V_{i,x,y}^j = \frac{\partial \left( \sum_{x,y} \textrm{img}_{x,y}^j\right)}{\partial c_i^1}\sqrt{4\pi}
    \label{eq:visibility}
\end{equation}

The derivative $\frac{\partial \left( \sum_{x,y} \textrm{img}_{x,y}^j\right)}{\partial c_i^1}$ can be easily and efficiently calculated with existing differentiable GS renderers.

\paragraph{Coloring} 
Next, the visibility of a Gaussian $i$ from camera $j$ can be used to compute a visibility-weighted color average of the ground truth image $\widehat{\textrm{img}}_{x,y}^j$:% similar to Equation \ref{eq:visibility}: % TODO: evtl ist hier die analogie nicht ganz klar...

\begin{equation}
    \hat C_i^j = \frac{\sum_{x,y}V_{i,x,y}^j \widehat{\textrm{img}}_{x,y}^j}{\sum_{x,y}V_{i,x,y}^j}
    = \frac{\partial \left(\sum_{x,y}\textrm{img}_{x,y}^j \widehat{\textrm{img}}_{x,y}^j\right)}{\partial c_i^1}\frac{\sqrt{4\pi}}{V_i^j}
    \label{eq:color_average}
\end{equation}

Here, we plugged in the previous results from Equations \ref{eq:pixel_visibility} and \ref{eq:visibility} into $V_{i,x,y}^j$ in the numerator and denominator. 
Again, the derivative $\frac{\partial \left(\sum_{x,y}\textrm{img}_{x,y}^j \widehat{\textrm{img}}_{x,y}^j\right)}{\partial c_i^1}$ can be easily obtained using a differentiable GS renderer.

If we neglect view-dependent effects (i.e. set $L=0$), $V_i^j$ and $\hat C_i^j$ already suffice to compute a visibility weighted color average $c_i^1$ for a Gaussian $i$ over all $N_c$ cameras that can be used e.g. to colorize diffuse surfaces, project view-independent neural features or to segment a scene:
\begin{equation}
    c_i^1 = \frac{\sum_{x,y,j} V_{i,x,y}^j \widehat{\textrm{img}}_{x,y}^j}{\sum_{x,y,j}V_{i,x,y}^j} = \frac{\sum_{j=1}^{N_c} V_i^j \hat C_i^j}{\sum_{j=1}^{N_c} V_i^j}
\end{equation}

% spherical harmonics
To incorporate view dependent effects, however, we also have to incorporate higher order spherical harmonics. To start, we need an easy and efficient way to compute $Y_m(\vec d)$. Again, we can make use of the chain rule to derive an expression for $Y_m(\vec d)$ as follows:
\begin{align}
    &\frac{\partial \left( \sum_{x,y} \textrm{img}_{x,y}^j\right)}{\partial c_i^m} = \underbrace{\frac{\partial \left( \sum_{x,y} \textrm{img}_{x,y}^j\right)}{\partial C_i(\vec d)}}_{=V_i^j} \underbrace{\frac{\partial C_i(\vec d)}{\partial c_i^m}}_{=Y_m(\vec d)}\\
    \Rightarrow & Y_m(\vec d) = \frac{\partial \left( \sum_{x,y} \textrm{img}_{x,y}^j\right)}{\partial c_i^m} \frac{1}{V_i^j} =: Y_{m,i}^j
    \label{eq:Y_comp}
\end{align}
Since $\vec d$ is the view direction pointing from camera $j$ to Gaussian $i$, we introduce an indexed notation $Y_{m,i}^j:= Y_m(\vec d)$. 
Here, the derivatives $\frac{\partial \left( \sum_{x,y} \textrm{img}_{x,y}^j\right)}{\partial c_i^m}$ 
can be obtained with little computational overhead since the derivatives of $\left(\sum_{x,y} \textrm{img}_{x,y}^j\right)$ need to be computed anyway to obtain the splat visibilities $V_i^j$ (see \cref{eq:visibility}).

% Least square problem (per Gaussian)
Now, to optimize view dependent effects, we aim to minimize the visibility-weighted squared differences between the ground truth images $\widehat{\textrm{img}}_{x,y}^j$ and the view dependent Gaussian colors $C_i(\vec d)$:% in viewing direction $\vec d$:

\begin{align}
    c_i^m=&\argmin_{c_i^m} \sum_{x,y,j} V_{i,x,y}^j\left( \widehat{\textrm{img}}_{x,y}^j - C_i(\vec d) \right)^2\\
    =&\argmin_{c_i^m} \sum_{x,y,j} V_{i,x,y}^j\left( \widehat{\textrm{img}}_{x,y}^j - \sum_{m}c_i^m Y_{m,i}^j \right)^2
    \label{eq:LS_full_sum_step_2}\\
    =&\argmin_{c_i^m} \sum_{j} V_{i}^j\left( \hat C_i^j - \sum_{m}c_i^m Y_{m,i}^j \right)^2
    \label{eq:LS_full_sum}
\end{align}

To obtain \Cref{eq:LS_full_sum_step_2}, we first plug in \Cref{eq:SH_coeff} into $C_i(\vec d)$ and use $Y_{m,i}^j$ from \Cref{eq:Y_comp}. Next, we decompose the weighted sum of squares using the visibility-weighted color averages $\hat C_i^j$ obtained from \Cref{eq:color_average} to derive \Cref{eq:LS_full_sum}. 
% Normal equation
This is a classical weighted least squares problem. 
To simplify the notation, in the following, we look at every Gaussian $i$ individually and introduce the diagonal matrix $\mathbf{V}_i\in \mathbb{R}^{N_c\times N_c}$ containing $(V_i^j)_{j=1,...,N_c}$ on its diagonal, the matrix $\mathbf{Y}_i \in \mathbb{R}^{N_c \times (L+1)^2}$ containing $(Y_{m,i}^j)_{j=1,...,N_c}^{m=1,...,(L+1)^2}$, the vector $\hat{\mathbf{C}}_i\in \mathbb{R}^{N_c}$ containing $(\hat C_i^j)_{j=1,...,N_c}$ and the vector $\mathbf{c}_i\in \mathbb{R}^{(L+1)^2}$ containing $(c_i^m)_{m=1,...,(L+1)^2}$ that we want to solve for. In this notation, \Cref{eq:LS_full_sum} can be rewritten in terms of a vector norm and solved by the normal equation:
\begin{equation}
\resizebox{\columnwidth}{!}{$
    \mathbf{c}_i = \argmin_{\mathbf{c}_i} ||\mathbf{V}_i^{\frac{1}{2}}(\hat{\mathbf{C}}_i-\mathbf{Y}_i\mathbf{c}_i)||^2\\
    = (\mathbf{Y}_i^\top\mathbf{V}_i\mathbf{Y}_i)^{-1}\mathbf{Y}_i^\top\mathbf{V}_i\hat{\mathbf{C}_i}
    \label{eq:normal_Eq}
$}
\end{equation}

This corresponds to a small linear system of $(L+1)^2$ equations for every Gaussian that can be solved quickly in parallel on a GPU. %  $(L+1)^2$ => typical values for L=3

% Regularization
\paragraph{Regularization}
If a Gaussian splat $i$ is seen only by a few cameras, \Cref{eq:normal_Eq} can become under-determined or ill-posed, resulting in a %singular, 
non-invertible Matrix $(\mathbf{Y}_i^\top\mathbf{V}_i\mathbf{Y}_i)$ or strong visual artifacts. Thus, we apply Tikhonov regularization \cite{tikhonov1977solutions} to \Cref{eq:normal_Eq}: 

\begin{equation}
    \mathbf{c}_i =(\mathbf{Y}_i^\top\mathbf{V}_i\mathbf{Y}_i+\mathbf{w}_i\mathbf{\Lambda})^{-1}\mathbf{Y}_i^\top\mathbf{V}_i\hat{\mathbf{C}_i}
    \label{eq:normal_Eq_regularized}
\end{equation}
Here, $\mathbf{\Lambda}$ is a diagonal matrix containing regularization weights $\lambda_m$ for the different SH coefficients and $\mathbf{w}_i=\sum_j V_i^j$ denotes the total visibility of a Gaussian $i$ over all cameras. 
Solving this regularized \Cref{eq:normal_Eq_regularized} corresponds to extending the minimization objective in \Cref{eq:LS_full_sum} by an extra penalty for high SH coefficients $c_i^m$ weighted by $\lambda_m$:

\begin{equation}
\resizebox{\columnwidth}{!}{$
    \argmin_{c_i^m} \sum_{j} V_{i}^j\left(\left( \hat C_i^j - \sum_{m}c_i^m Y_{m,i}^j \right)^2 + \sum_m \lambda_m (c_i^m)^2\right)
    \label{eq:LS_full_sum_regularized}
$}
\end{equation}

% Iterative refinement
\paragraph{Iterative Refinement}
So far, all Gaussian splats are colorized individually. Under the assumption that all Gaussians are rendered disjointly, this would lead to correct rendering results. 
In practice, however, most Gaussians are transparent, which allows colors from different splats to superimpose and significantly impacts the final rendering. % TODO: Evtl bsp visualisierung zeigen
To also cover such color-mixing interactions between Gaussians, we propose an iterative refinement scheme: 

As shown in lines 8-10 of \cref{alg:colorization}, each refinement step works similarly to the initial colorization procedure (lines 1-6). 
However, instead of directly using the ground truth images ($\widehat{\textrm{img}}_{x,y}^j$ in line 4) to compute the target colors $\hat C_i^j$, the difference between the ground truth images and the rendered images ($\widehat{\textrm{img}}_{x,y}^j-\textrm{img}_{x,y}^j\textrm{.detach()}$ in line 9) is used to update the SH coefficients in line 10. %\todo{evtl könnte das noch besser formuliert werden}
Note, that this refinement process only requires recomputing $\hat C_i^j$ since the geometry (and thus $V_i^j$, $Y_{m,i}^j$ and $\mathbf{w}_i$) remains fixed. Furthermore, since $(\mathbf{Y}_i^\top\mathbf{V}_i\mathbf{Y}_i+\mathbf{w}_i\mathbf{\Lambda})^{-1}$ stays the same for all refinement steps and all color channels, we compute the inverse only once and reuse it. 
A more detailed derivation of the refinement steps is provided in our supplementary material.
% TODO: more details on regularized refinement steps in appendix

\begin{algorithm}
\caption{Colorization algorithm}
\label{alg:colorization}
\KwData{GT images: $\widehat{\textrm{img}}_{x,y}^j$, 
camera poses: $P_j$, regularization weights: $\lambda_m$, \\
\hspace{0.9cm} Gaussian means and covariances: $\mu_i$, $\Sigma_i$}
\KwResult{Gaussian SH coefficients: $c_i^m$}
\BlankLine

$\textrm{img}_{x,y}^j \leftarrow \textrm{render}(P_j,(\vec \mu_i,\Sigma_i,c_i^m))$

$V_i^j \leftarrow \frac{\partial \left( \sum_{x,y} \textrm{img}_{x,y}^j\right)}{\partial c_i^1}\sqrt{4\pi}$

$Y_{m,i}^j \leftarrow \frac{\partial \left( \sum_{x,y} \textrm{img}_{x,y}^j\right)}{\partial c_i^m} \frac{1}{V_i^j}$

$\hat C_i^j \leftarrow \frac{\partial \left(\sum_{x,y}\textrm{img}_{x,y}^j \widehat{\textrm{img}}_{x,y}^j\right)}{\partial c_i^1}\frac{\sqrt{4\pi}}{V_i^j}$

$\mathbf{w}_i \leftarrow \sum_j V_i^j$

$\mathbf{c}_i \leftarrow(\mathbf{Y}_i^\top\mathbf{V}_i\mathbf{Y}_i+\mathbf{w}_i\mathbf{\Lambda})^{-1}\mathbf{Y}_i^\top\mathbf{V}_i\hat{\mathbf{C}_i}$

\For{$\textrm{step} \leftarrow 1$ \KwTo $N_\textrm{refine}$}{
    
    $\textrm{img}_{x,y}^j \leftarrow \textrm{render}(P_j,(\vec \mu_i,\Sigma_i,c_i^m))$

    $\hat C_i^j \leftarrow \frac{\partial \left(\sum_{x,y}\textrm{img}_{x,y}^j \left(\widehat{\textrm{img}}_{x,y}^j-\textrm{img}_{x,y}^j\textrm{.detach()}\right)\right)}{\partial c_i^1}\frac{\sqrt{4\pi}}{V_i^j}$

    $\mathbf{c}_i \leftarrow \mathbf{c}_i+(\mathbf{Y}_i^\top\mathbf{V}_i\mathbf{Y}_i+\mathbf{w}_i\mathbf{\Lambda})^{-1}(\mathbf{Y}_i^\top\mathbf{V}_i\hat{\mathbf{C}_i}-\mathbf{w}_i\mathbf{\Lambda}\mathbf{c}_i)$

}
\Return{$c_i^m$}%\;
\end{algorithm}
\section{Experiments}
\label{sec:experiments}

We evaluated the novel colorization algorithm in three different scenarios: First, we tested its performance on a scene relighting task. Second, we used the algorithm to enrich a 3D scene with neural features from a DINOv2 model and third, we applied the algorithm to do efficient 3D semantic scene segmentation by mapping segmentation masks provided by the SAM 2 model onto the Gaussian splats.

\paragraph{Implementation Details}

The recolorization algorithm is implemented in PyTorch based on the original Gaussian Splatting framework by Kerbl et al. \cite{kerbl3Dgaussians}
Following most works on Gaussian splatting, we chose the maximum SH order to be $L=3$. Furthermore, we empirically found the following regularization values for $\lambda_m$ worked out best: $\lambda_1=10^{-5};\lambda_{2-4}=10^{-4};\lambda_{5-9}=10^{-3};\lambda_{10-16}=10^{-2}$.

%Hardware description
All of our experiments were conducted on a machine based on a Intel Core i9-14900KF CPU with a NVIDIA GeForce RTX 4090 GPU and 64GB DDR5 RAM.
% 945497 Gaussian splats

\subsection{Relighting}
\label{sec:relighting}

Relighting is a challenging task, in particular when considering complex illumination effects such as sub-surface scattering, caustics or intricate shadows that go beyond conventional approximations such as e.g. the Blinn-Phong model. 
Thus, following research on light stages \cite{sun_light_stage_superres_2020,debevec2012light} and BTF capture devices \cite{rump10groudtruth,filip18evaluating}, we apply our instant colorization algorithm to map images of the same scene under different lighting conditions onto a common scene representation consisting of gaussian splats. 
Because relighting involves high-frequency details and strong view-dependent effects, it is a demanding test bench for texture mapping algorithms. 
While there exist already datasets for relighting, we couldn’t find an open dataset that captures static real-world scenes under various lighting conditions from sufficiently many camera angles to perform 3DGS.
Thus, we captured a new dataset (more details on the dataset are provided in the supplementary).

%One of the primary goals of this work is to enable fast and flexible relighting of 3D Gaussian Splat representations directly from multi-view image data. While traditional Gaussian Splatting approaches focus on photorealistic novel-view synthesis under fixed illumination, our method extends this concept by efficiently mapping color information from differently illuminated views back onto the underlying Gaussian splats. 
% TODO: das ist noch ungünstig formuliert... es gibt ja auch andere arbeiten für relighting. evtl sollte relighting auch nicht "primary goal"

\begin{figure}[t]
\centering
\captionsetup[subfigure]{font=footnotesize}
\setlength{\tabcolsep}{2pt}
\renewcommand{\arraystretch}{0.9}

\begin{tabular}{cc}

% -------- Row 1 --------
\begin{subfigure}[t]{0.47\linewidth}\centering
\includegraphics[height=0.13\textheight]
{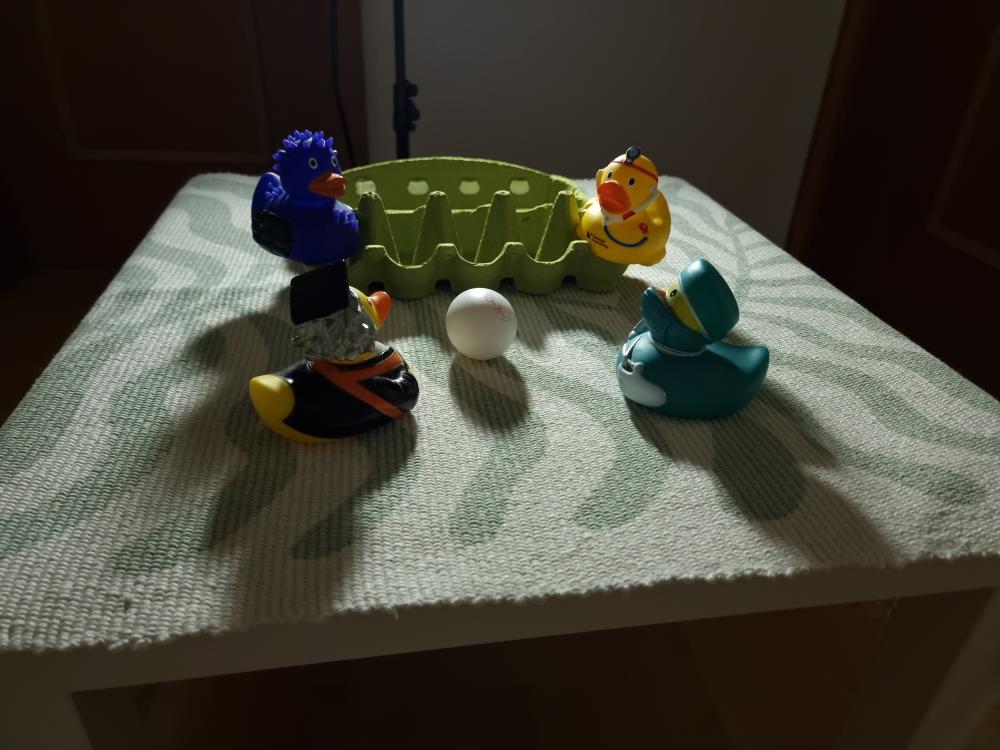}
\subcaption{Duck scene, light 1}
\end{subfigure} &
\begin{subfigure}[t]{0.47\linewidth}\centering
\includegraphics[height=0.13\textheight]
{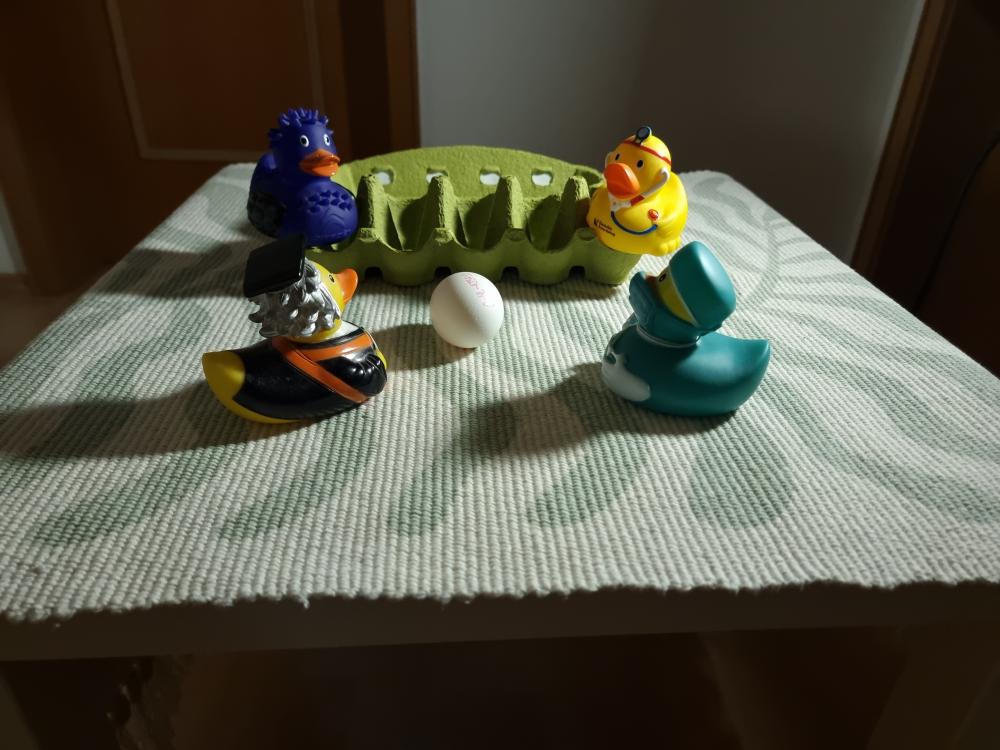}
\subcaption{Duck scene, light 2}
\end{subfigure}\\

% -------- Row 2 --------
\begin{subfigure}[t]{0.47\linewidth}\centering
\includegraphics[height=0.13\textheight]
{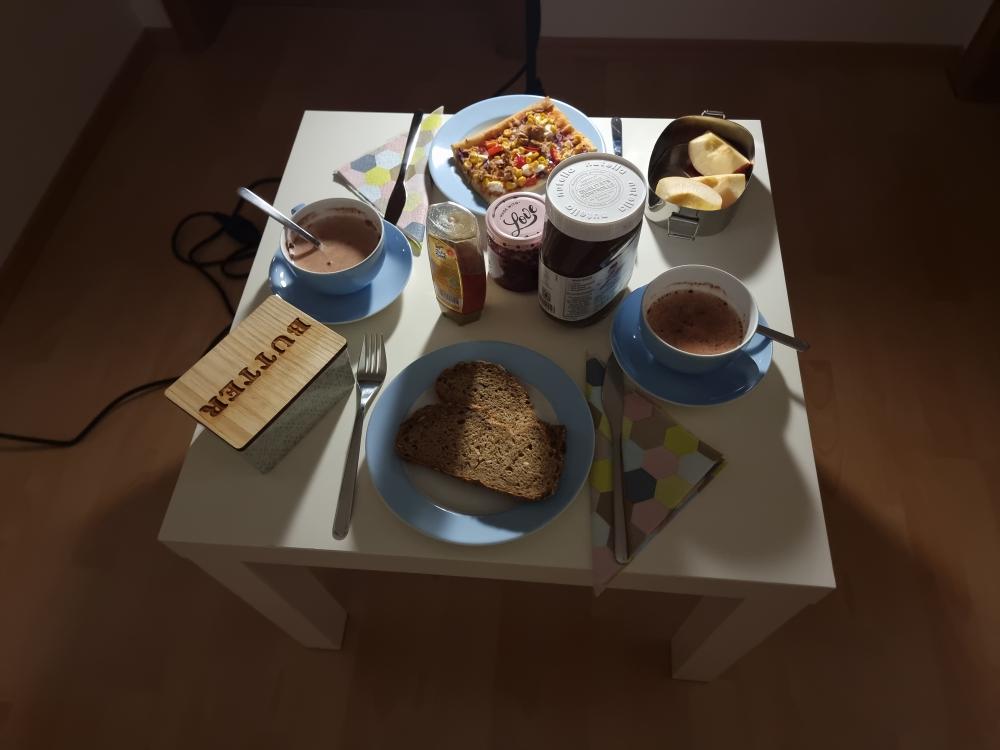}
\subcaption{Food scene, light 1}
\end{subfigure} &
\begin{subfigure}[t]{0.47\linewidth}\centering
\includegraphics[height=0.13\textheight]
{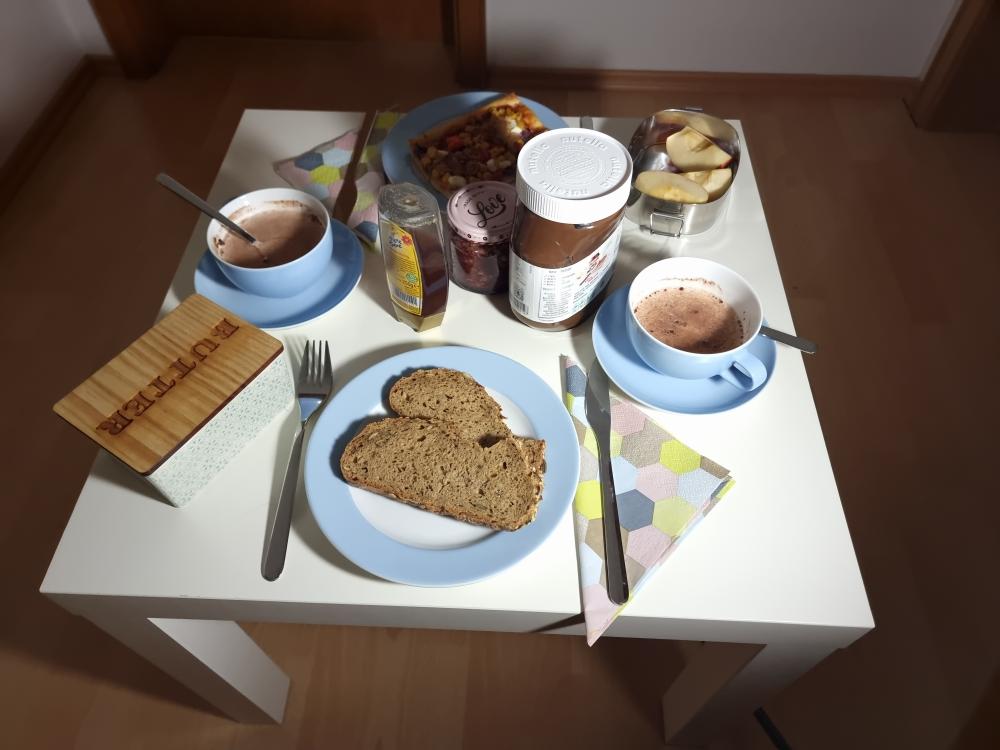}
\subcaption{Food scene, light 2}
\end{subfigure}\\

% -------- Row 3 --------
\begin{subfigure}[t]{0.47\linewidth}\centering
\includegraphics[height=0.13\textheight]
{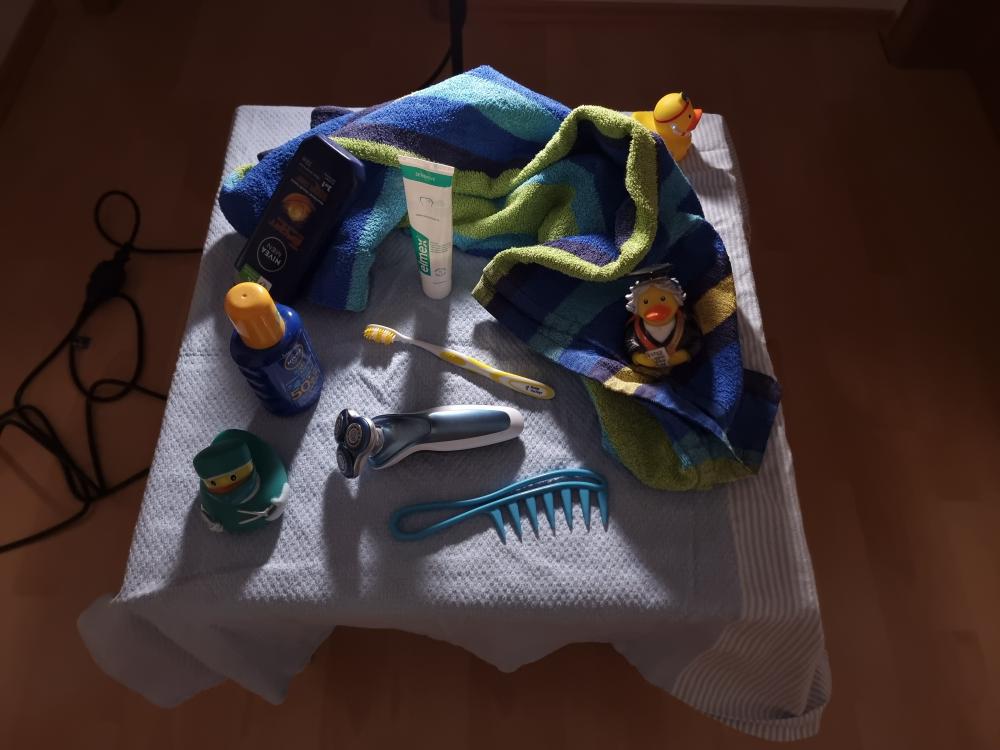}
\subcaption{Bathroom scene, light 1}
\end{subfigure}   &
\begin{subfigure}[t]{0.47\linewidth}\centering
\includegraphics[height=0.13\textheight]
{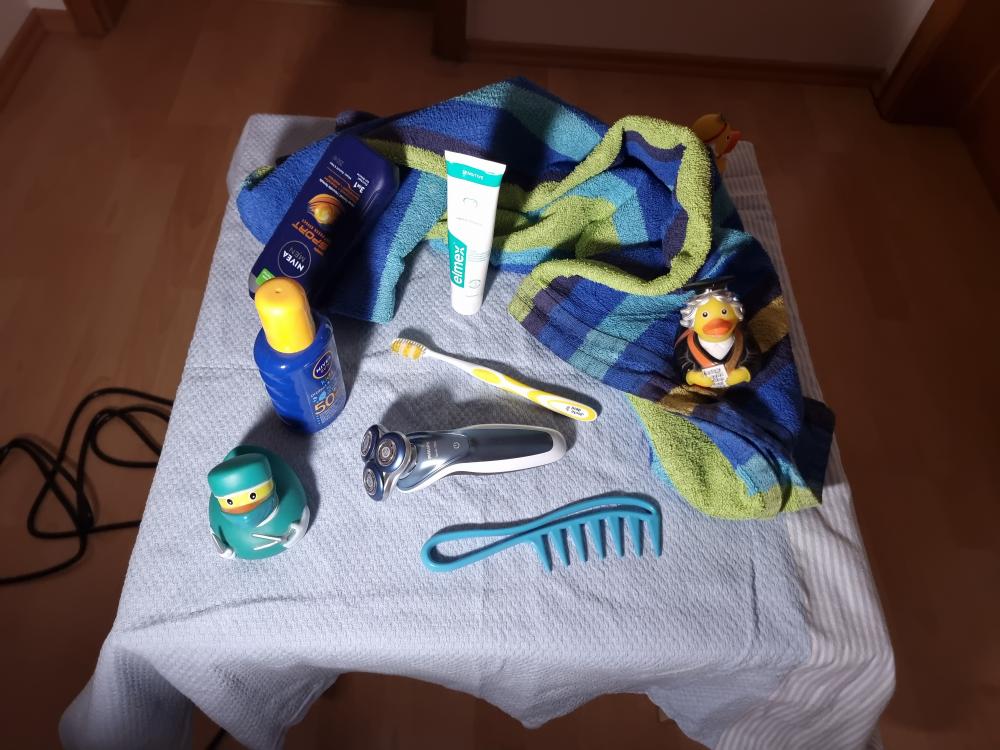}
\subcaption{Bathroom scene, light 2}
\end{subfigure}

\end{tabular}

\vspace{-2mm}
\caption{
Examples of our captured scenes under different lighting conditions.
a, b, c: light 1. d, e, f: light 2.
}
\label{fig:data_examples}
\end{figure}

\paragraph{Setup}
We captured several static scenes under different lighting conditions to provide a suitable testbed for relighting (see \cref{fig:data_examples}). 
Each scene was recorded multiple times with similar camera trajectories on a smartphone but with varying illumination setups, such as different light sources or light directions. 

\begin{figure}
     \centering
     \begin{subfigure}[b]{0.49\textwidth}
        \centering
        \includegraphics[width=\linewidth]{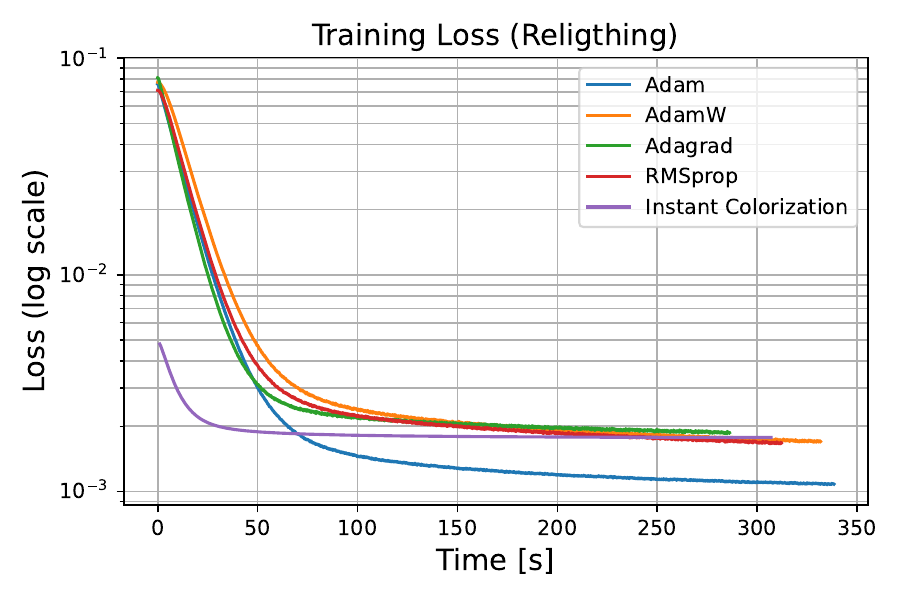}
        \caption{$L_2$ Training loss (log scale, lower is better) comparison of our instant baseline with Adam (lr=0.0025), AdamW (lr=0.0025), RMSprop (lr=0.0025), and Adagrad (lr=0.1) over time. 
        %\todo{Duck scene, Light 2 (corresponds light_index=1)}
        }
        \label{fig:loss_train_mit_reg}
     \end{subfigure}
     \hfill
     \begin{subfigure}[b]{0.49\textwidth}
        \centering
        \includegraphics[width=\linewidth]{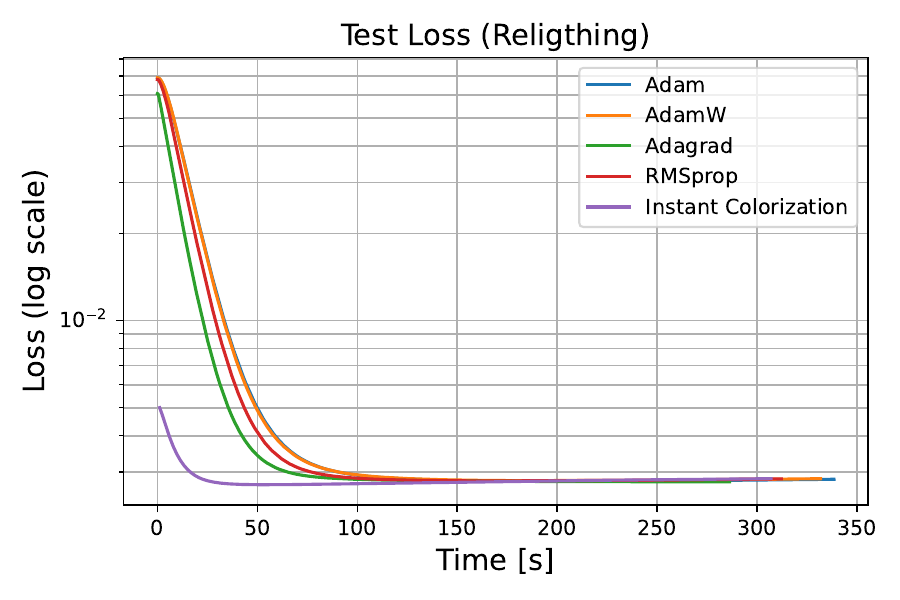}
        \caption{$L_2$ Test loss (log scale, lower is better) comparison of our instant baseline with Adam (lr=0.0025), AdamW (lr=0.0025), RMSprop (lr=0.0025), and Adagrad (lr=0.1) over time.
        }
        \label{fig:loss_test_mit_reg}
     \end{subfigure}
        \caption{Train and Test Loss curves for different optimizers. Our instant colorization approach achieves the lowest test loss in the shortest amount of time. Adam tends to overfit.}
        \label{fig:instant_colorization_curves}
\end{figure}

To ensure alignment of the captured photos with the Gaussian splat geometry that was reconstructed from the "original" lighting condition (see \cref{fig:architecture_overview}) we relied on COLMAP. 
For the subsequent colorization, we applied the proposed visibility-weighted least squares solver (see \cref{sec:method}) to map the images from different illumination conditions onto the Gaussian splats and compared its performance to gradient descent based alternatives like Adam, AdamW, Adagrad and RMSprop (see \Cref{fig:instant_colorization_curves}).

\begin{figure}[h!]
    \centering
    \captionsetup[subfigure]{font=footnotesize}
    \setlength{\tabcolsep}{2pt}
    \renewcommand{\arraystretch}{0.9}
    \begin{tabular}{cc}
    \begin{subfigure}[t]{0.48\linewidth}
        \centering
        \includegraphics[width=\linewidth]{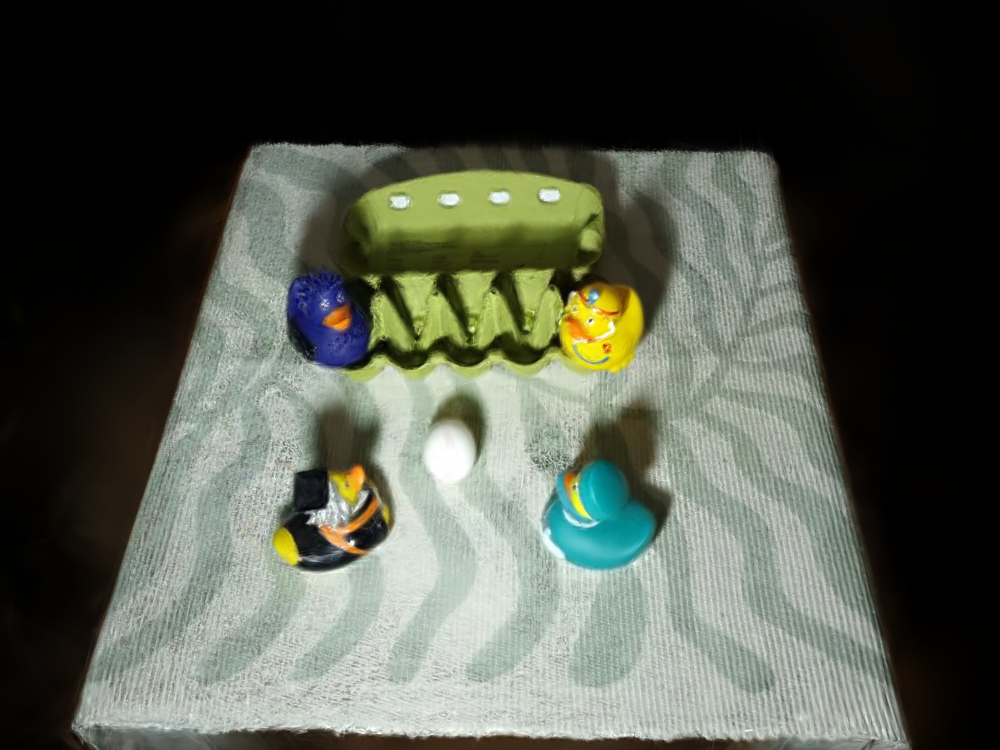}
        \caption{Duck scene, relighted}    
    \end{subfigure}&
    \begin{subfigure}[t]{0.48\linewidth}
        \centering
        \includegraphics[width=\linewidth]{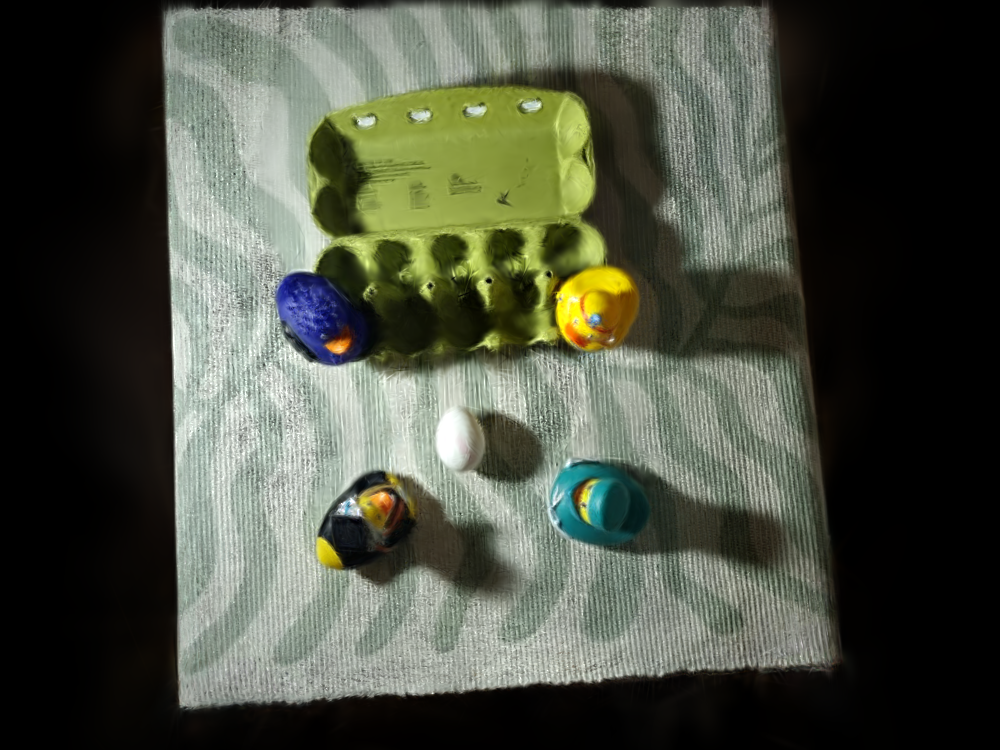}
        \caption{Duck scene, relighted}
    \end{subfigure}\\

    %\vspace{0.5em}
    
    \begin{subfigure}[t]{0.48\linewidth}
        \centering
        \includegraphics[width=\linewidth]{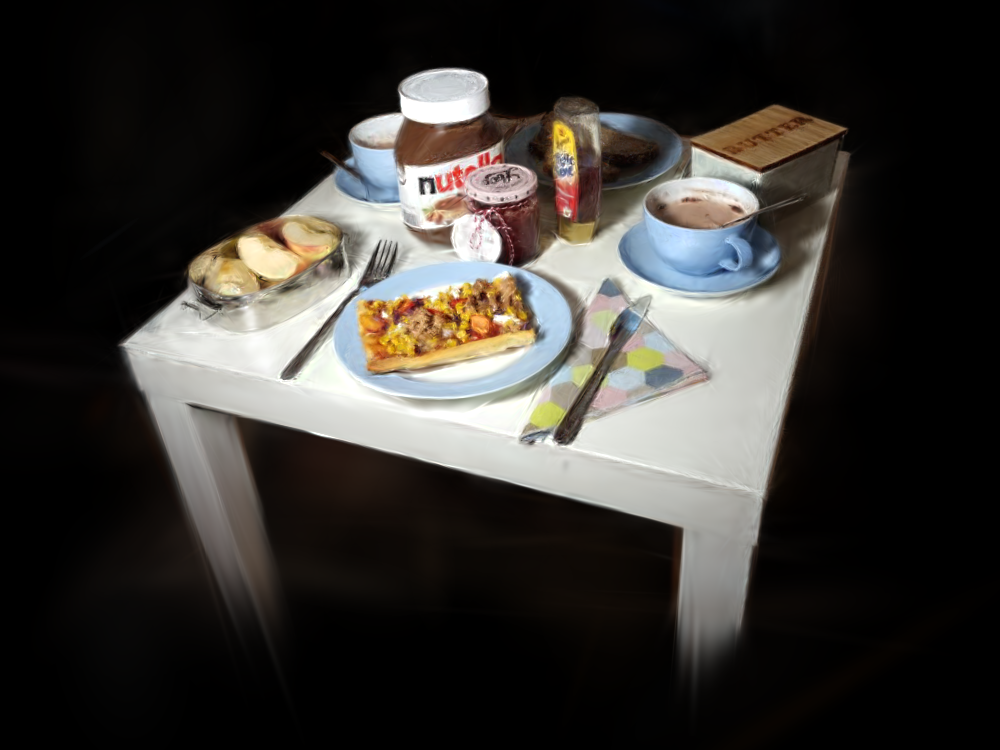}
        \caption{Food scene, relighted}    
    \end{subfigure}&
    \begin{subfigure}[t]{0.48\linewidth}
        \centering
        \includegraphics[width=\linewidth]{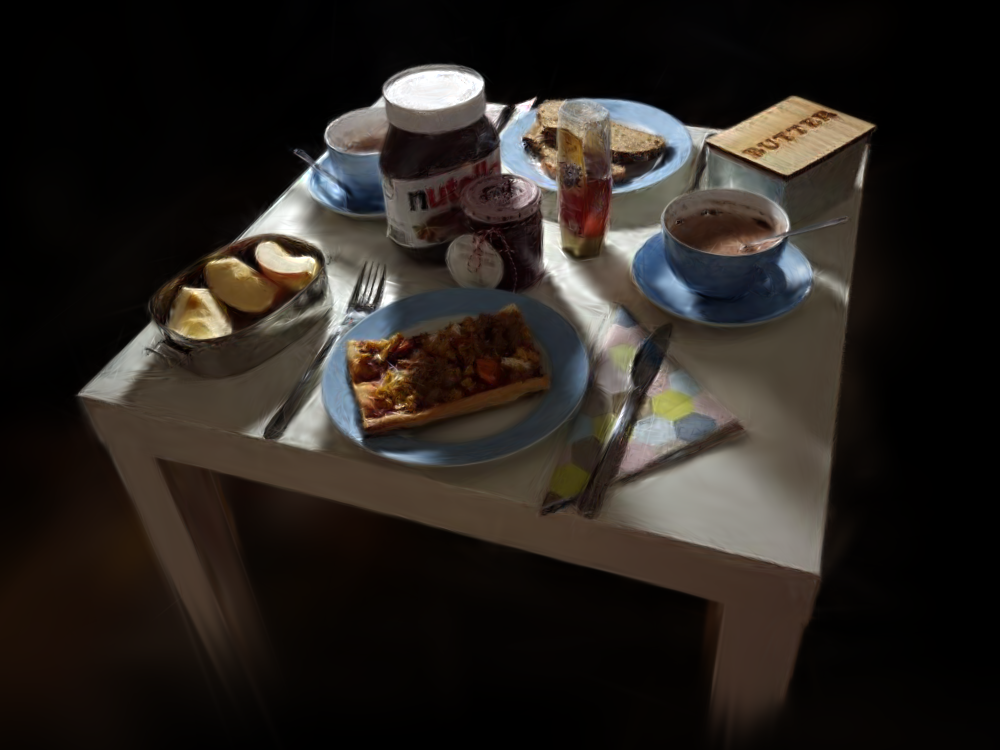}
        \caption{Food scene, relighted}
    \end{subfigure}\\

    %\vspace{0.5em}
    
    \begin{subfigure}[t]{0.48\linewidth}
        \centering
        \includegraphics[width=\linewidth]{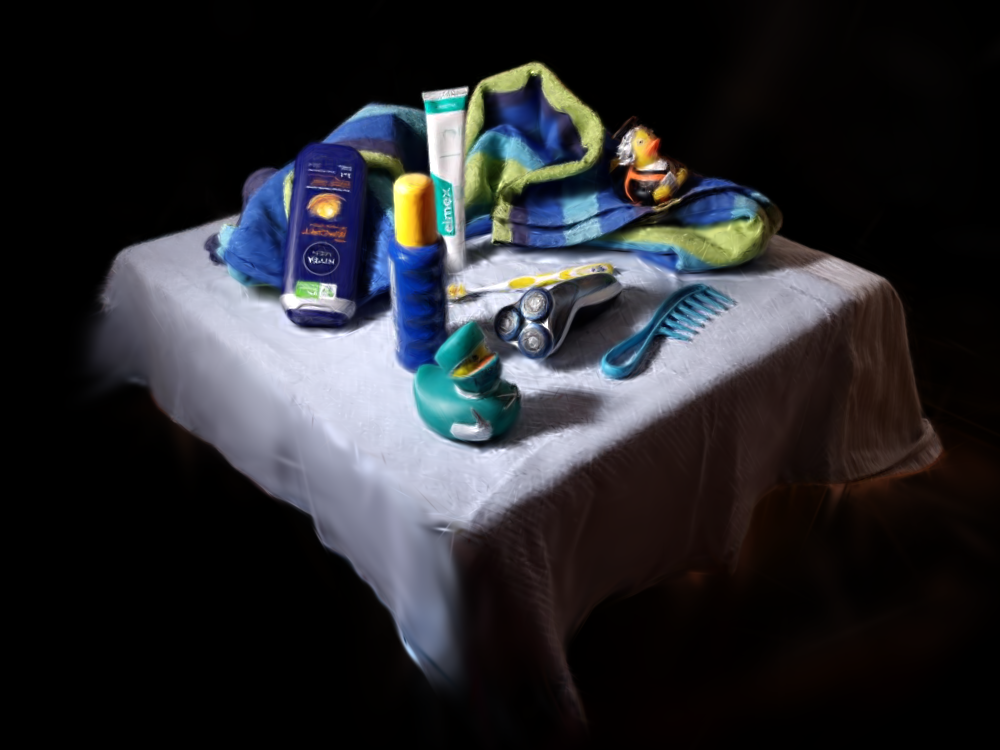}
        \caption{Bathroom scene, relighted}
    \end{subfigure}&
    \begin{subfigure}[t]{0.48\linewidth}
        \centering
        \includegraphics[width=\linewidth]{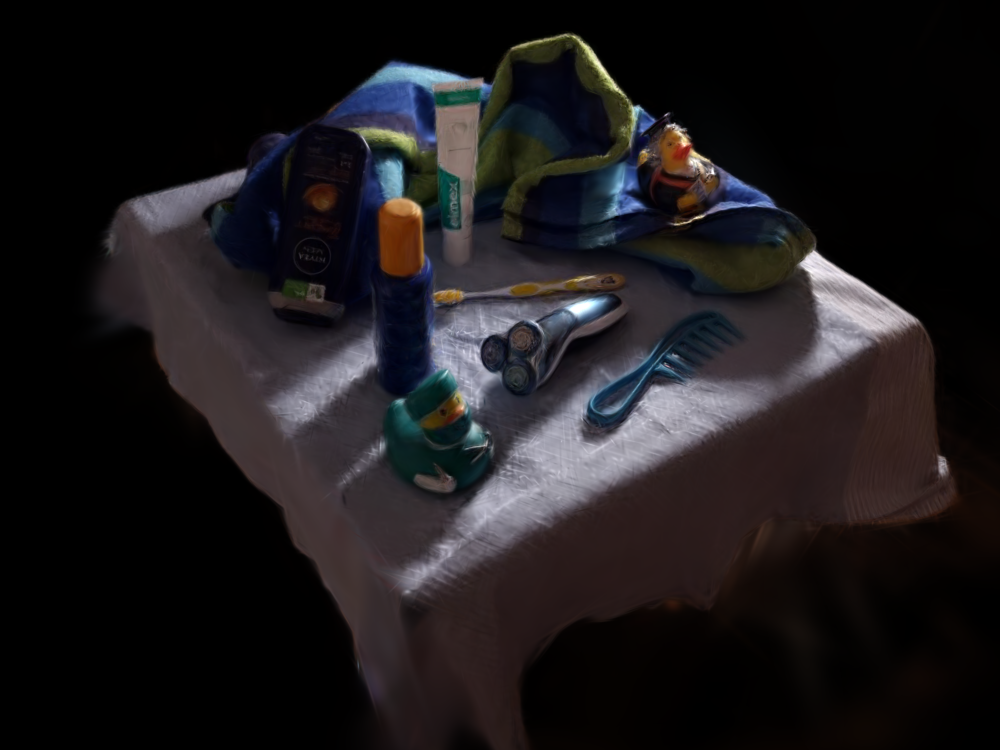}
        \caption{Bathroom scene, relighted}    
    \end{subfigure}

    \end{tabular}
    
    \vspace{-2mm}
    \caption{
    Qualitative results for 3 scenes with different relightings by the instant colorization algorithm.
    }
    \label{fig:relighting_qualitative}
\end{figure}

\begin{figure}[!h]
\centering
\captionsetup[subfigure]{font=footnotesize}
\setlength{\tabcolsep}{2pt}
\renewcommand{\arraystretch}{0.9}

\begin{tabular}{cc}

    \begin{subfigure}[t]{0.48\linewidth}
        \centering
        \includegraphics[width=\linewidth]{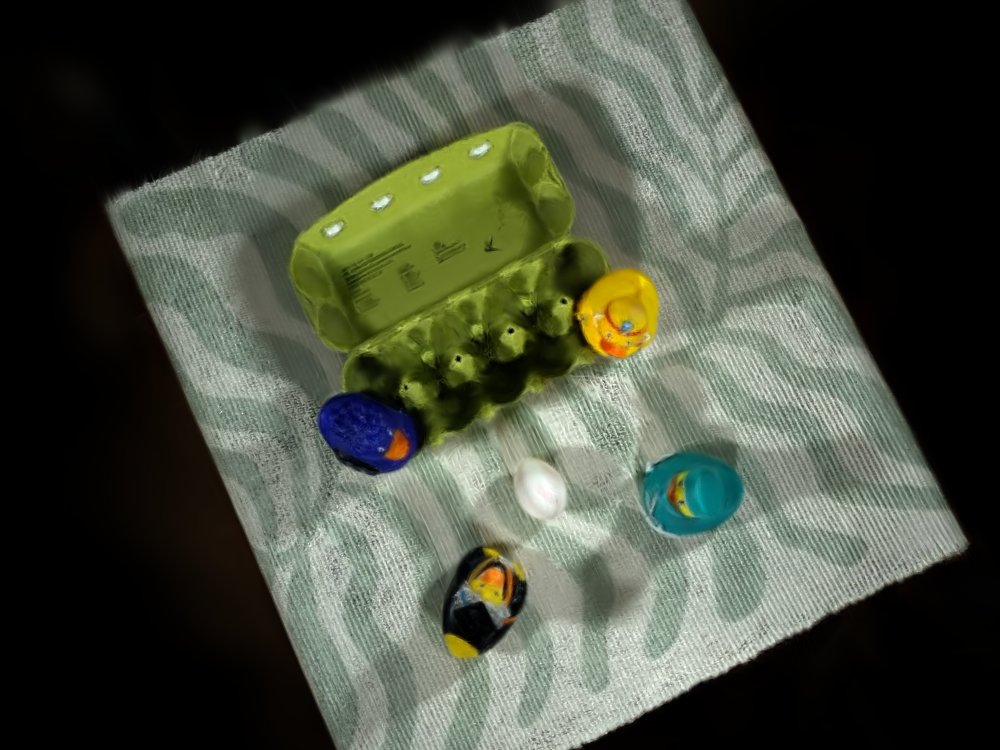}
        \caption{Duck, combined relighting.}
    \end{subfigure}&
    \begin{subfigure}[t]{0.48\linewidth}
        \centering
        \includegraphics[width=\linewidth]{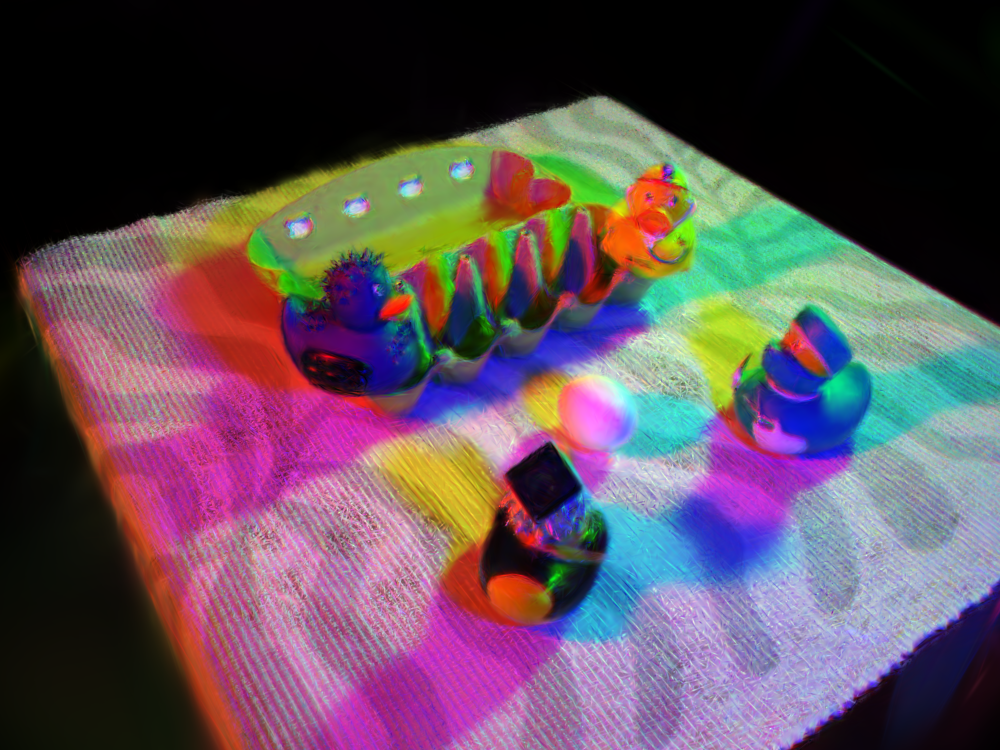}
        \caption{Duck, colored relighting.}    
    \end{subfigure}\\
    
    \begin{subfigure}[t]{0.48\linewidth}
        \centering
        \includegraphics[width=\linewidth]{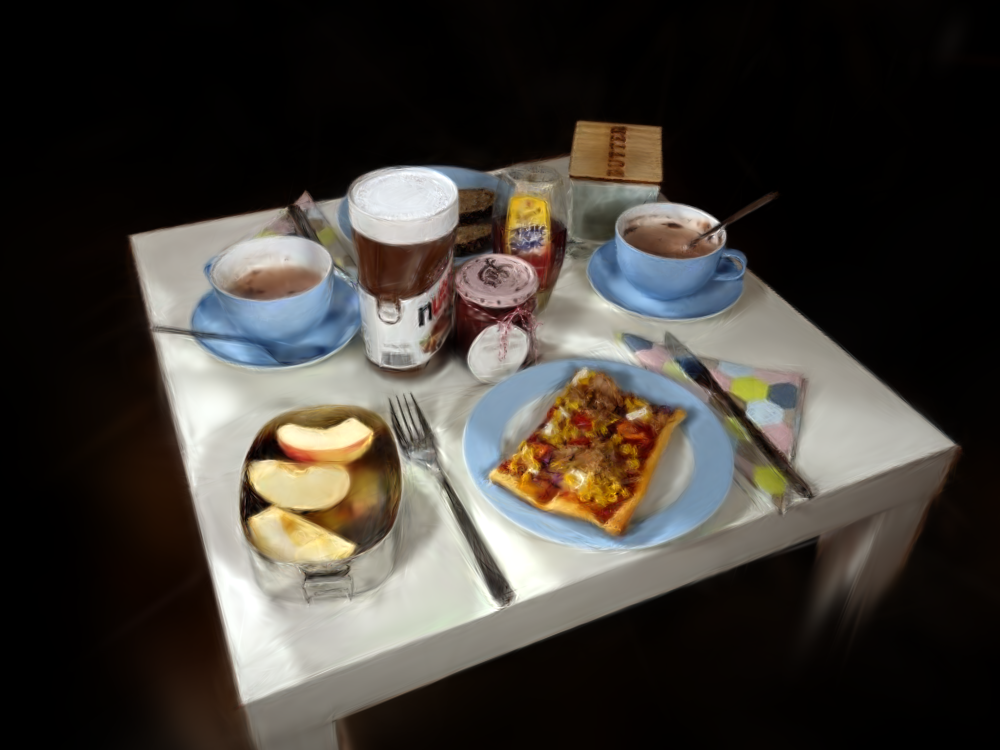}
        \caption{Food, combined relighting.}    
    \end{subfigure}&
    \begin{subfigure}[t]{0.48\linewidth}
        \centering
        \includegraphics[width=\linewidth]{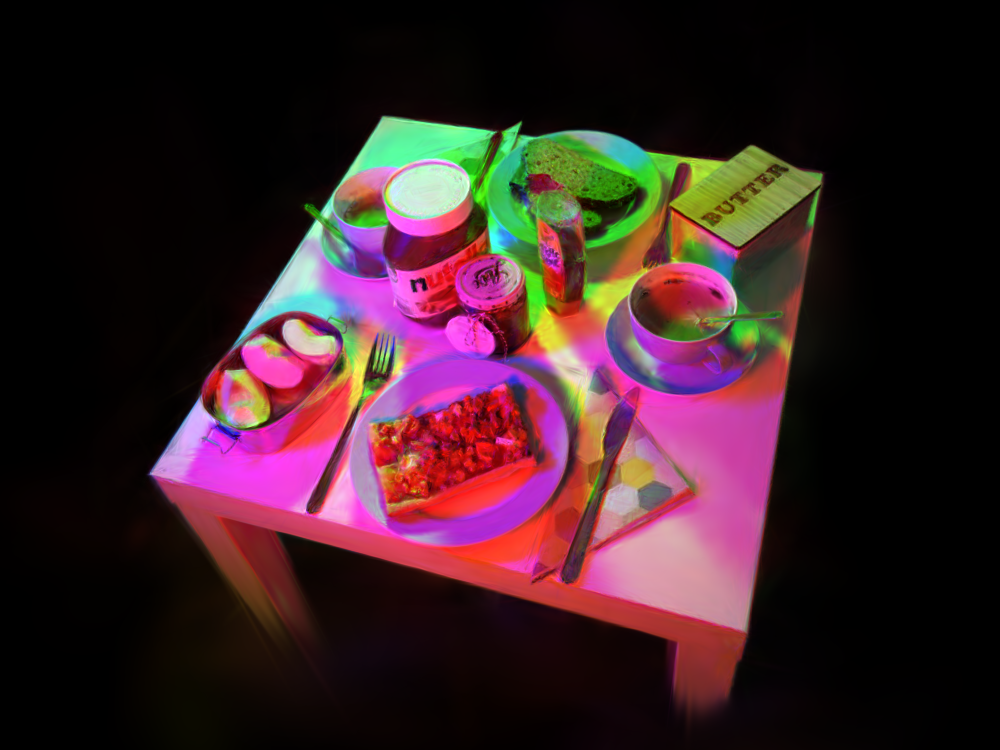}
        \caption{Food, colored relighting.}
    \end{subfigure}\\
    
    \begin{subfigure}[t]{0.48\linewidth}
        \centering
        \includegraphics[width=\linewidth]{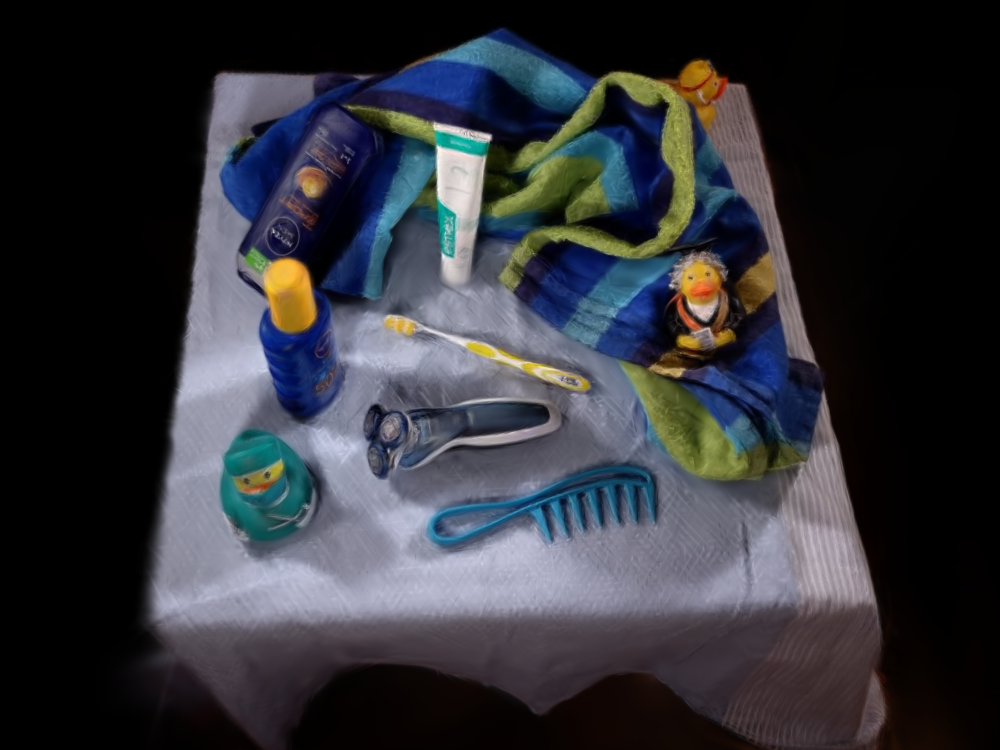}
        \caption{Bathroom, combined relighting.}
    \end{subfigure}&
    \begin{subfigure}[t]{0.48\linewidth}
        \centering
        \includegraphics[width=\linewidth]{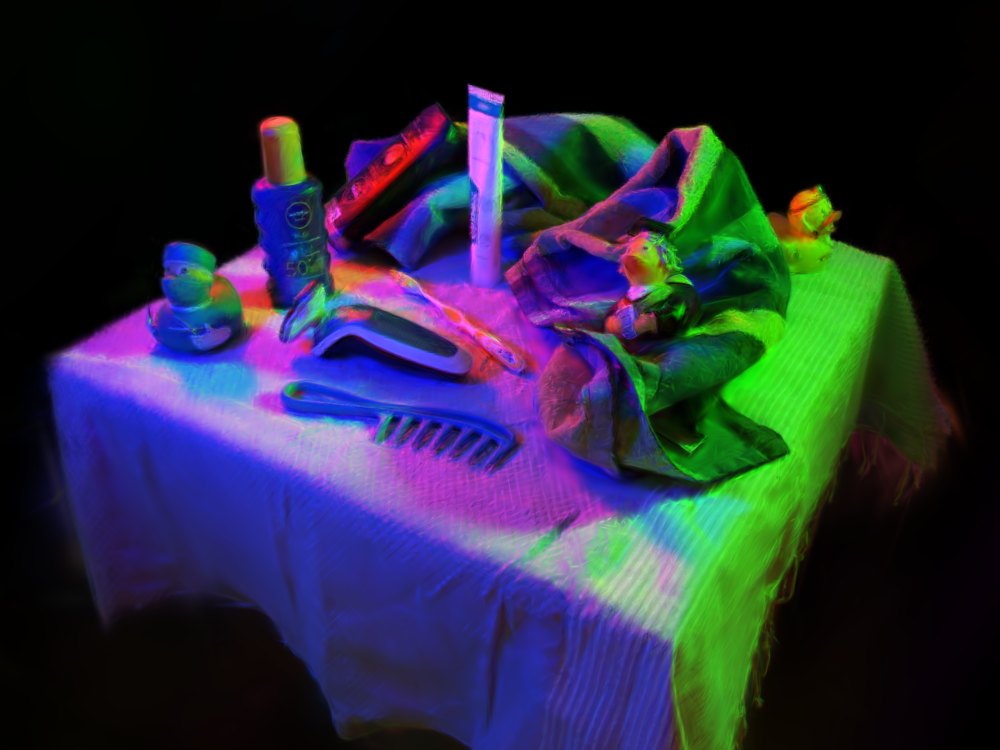}
        \caption{Bathroom, colored relighting.}    
    \end{subfigure}

\end{tabular}

\vspace{-2mm}
    \caption{Light fields can be linearly combined. (left): equal-weight combination of three lighting conditions. (right): same combinations with additional RGB color filters applied.
    }  
    \label{fig:combined_relighting}
\end{figure}

\paragraph{Qualitative results}
\Cref{fig:relighting_qualitative} shows accurate scene appearances for different lighting variations mapped by our colorization algorithm. % while maintaining smooth transitions across views.
Since light fields are linear, the color (SH) coefficients of the gaussian splats from different colorizations can be combined (and filtered) to create novel scene lightings that were not contained in the dataset (see \cref{fig:combined_relighting}).

\begin{figure}[h!]
    \centering
    \includegraphics[width=\linewidth]{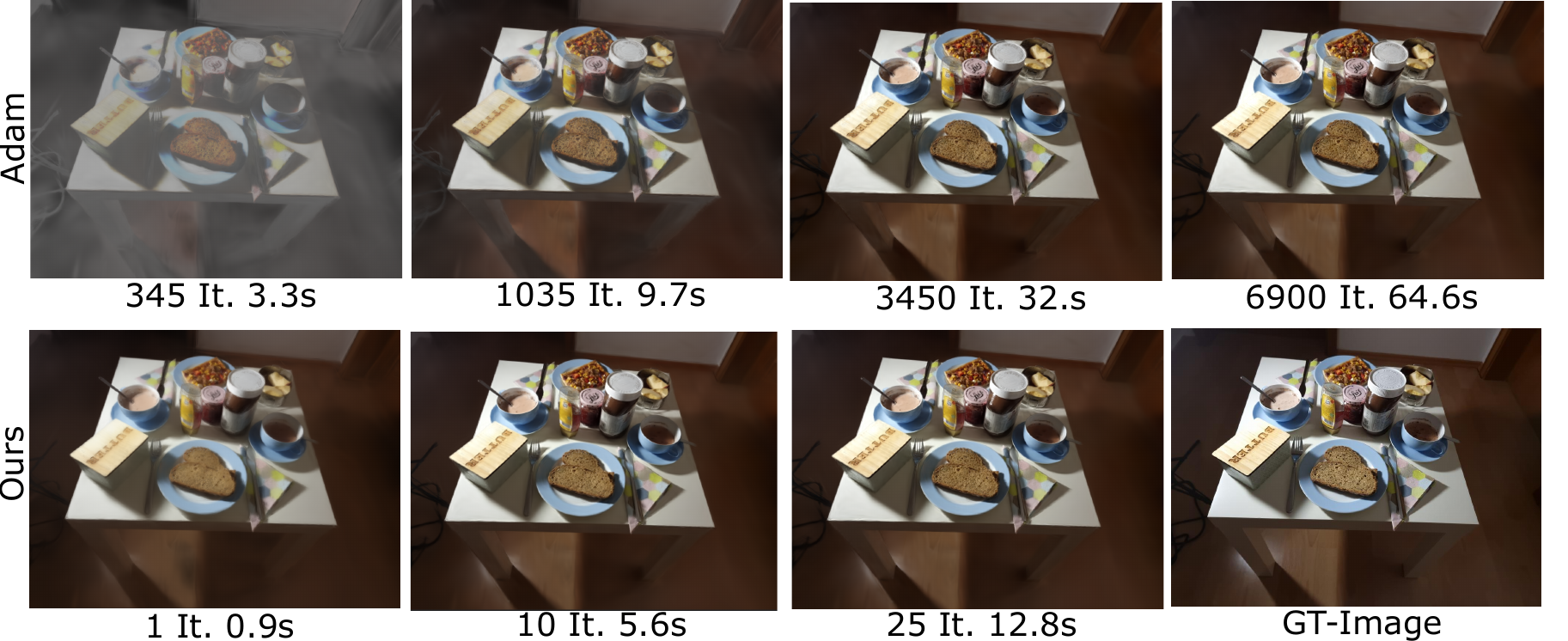}
    \caption{Top row: Results using ADAM optimization show gradual improvement but require significantly more iterations and time to achieve high-quality appearance.
    Bottom row: The instant colorization algorithm converges much faster, achieving high quality in just a few iterations, or seconds.
    }
    \label{fig:relighting_train_qualitative}
\end{figure}

\paragraph{Quantitative results}
As can be seen in Table \ref{tab:test_metrics} and \Cref{fig:instant_colorization_curves}, our instant colorization approach achieves better test metrics while being up to an order of magnitude faster than gradient descent based alternatives like Adam \cite{kingma2017adammethodstochasticoptimization}, AdamW \cite{Loshchilov_AdamW}, Adagrad or RMSprop. 
Already after the first iteration (around 1s), our method achieves lower test losses than gradient descent based alternatives after 40-50s. 
This is also reflected in the perceived reconstruction quality over time as shown in \Cref{fig:relighting_train_qualitative} %\todo{Tabelle!}
and demonstrates that solving the color mapping analytically through the normal equation provides a highly efficient and stable alternative to gradient-descent based relighting.

\begin{table}
  \caption{Test metric comparison of the instant colorization approach to established gradient descent methods}
  \label{tab:test_metrics}
  \centering
  \begin{tabular}{@{}l|c|c|c@{}}
    \toprule
    Duck / Light 1 &  $L_1 (\downarrow)$ & $L_2 (\downarrow)$ & PSNR $(\uparrow)$\\
    \midrule
    Adam  & $3.379{\times}10^{-2}$ & $2.777{\times}10^{-3}$ & $25.98$ \\
    \midrule
    AdamW  & $3.394{\times}10^{-2}$ & $2.794{\times}10^{-3}$ & $25.95$ \\
    \midrule
    RMSprop  & $3.387{\times}10^{-2}$ & $2.790{\times}10^{-3}$ & $25.96$ \\
    \midrule
    Adagrad  & $3.366{\times}10^{-2}$ & $2.775{\times}10^{-3}$ & $26.01$ \\
    %\midrule
    %Ours (w/o reg) & $3.4657{\times}10^{-2}$ & $2.924{\times}10^{-3}$ & $25.75$ \\
    \midrule
    \textbf{Ours} & $\mathbf{3.324{\times}10^{-2}}$ & $\mathbf{2.709{\times}10^{-3}}$ & $\mathbf{26.08}$ \\
    \bottomrule
  \end{tabular}
\end{table}

\iffalse
\begin{figure}[t]
\centering
\setlength{\tabcolsep}{2pt}
\renewcommand{\arraystretch}{0.9}

\begin{tabular}{ccc}
\begin{subfigure}[t]{0.31\linewidth}\centering
  \includegraphics[height=0.13\textheight]{figures/tisch_original.jpg}
  \subcaption*{(a) Input (LC1)}
\end{subfigure} &
\begin{subfigure}[t]{0.31\linewidth}\centering
  \includegraphics[height=0.13\textheight]{figures/tisch_set2_original.jpg}
  \subcaption*{(b) Input (LC2)}
\end{subfigure} &
\begin{subfigure}[t]{0.31\linewidth}\centering
  \includegraphics[height=0.13\textheight]{figures/tisch_set3_original.jpg}
  \subcaption*{(c) Input (LC3)}
\end{subfigure} \\

\begin{subfigure}[t]{0.31\linewidth}\centering
  \includegraphics[height=0.13\textheight]{figures/instant_iterativ_set1.png}
  \subcaption*{(d) Relighted (LC1)}
\end{subfigure} &
\begin{subfigure}[t]{0.31\linewidth}\centering
  \includegraphics[height=0.13\textheight]{figures/instant_iterativ_set2.png}
  \subcaption*{(e) Relighted (LC2)}
\end{subfigure} &
\begin{subfigure}[t]{0.31\linewidth}\centering
  \includegraphics[height=0.13\textheight]{figures/instant_iterativ_set3.png}
  \subcaption*{(f) Relighted (LC3)}
\end{subfigure}
\end{tabular}

\vspace{-2mm}
\caption{Relighting under three lighting conditions. Top: input images. Bottom: relighted results. (LC = light condition)}
\label{fig:relighting_qualitative_table}
\end{figure}
\fi

\subsection{Feature Enrichment}

Beyond relighting, dense semantic feature maps can also be projected into a geometrically consistent 3D representation to enrich an existing scene geometry with feature embeddings. % gut
By aligning spatial information of latent feature spaces used e.g. in vision or vision-language models, such enriched representations are an important backbone of countless downstream applications such as making scenes queryable by multimodal data \cite{conceptfusion2023, clipfields2023,lerf2023, peng2023openscene, maggio2024clio, garfield2024}.%, guenther2026sgb, igelbrink2026disc}.%, most prominently natural language \cite{lerf2023, peng2023openscene, maggio2024clio, garfield2024, guenther2026sgb, igelbrink2026disc}.

%As a result, the scene representation becomes queryable by multimodal data, most prominently natural language \cite{lerf2023, peng2023openscene, maggio2024clio, garfield2024, guenther2026sgb, igelbrink2026disc}. % vllt besser in realted work?
%Other sensor modalities can also be used if they can be meaningfully transformed into the same latent space and compared with the features of the map \cite{conceptfusion2023, clipfields2023}.

\begin{figure}[h!]
  \centering
  \includegraphics[width=1\linewidth]{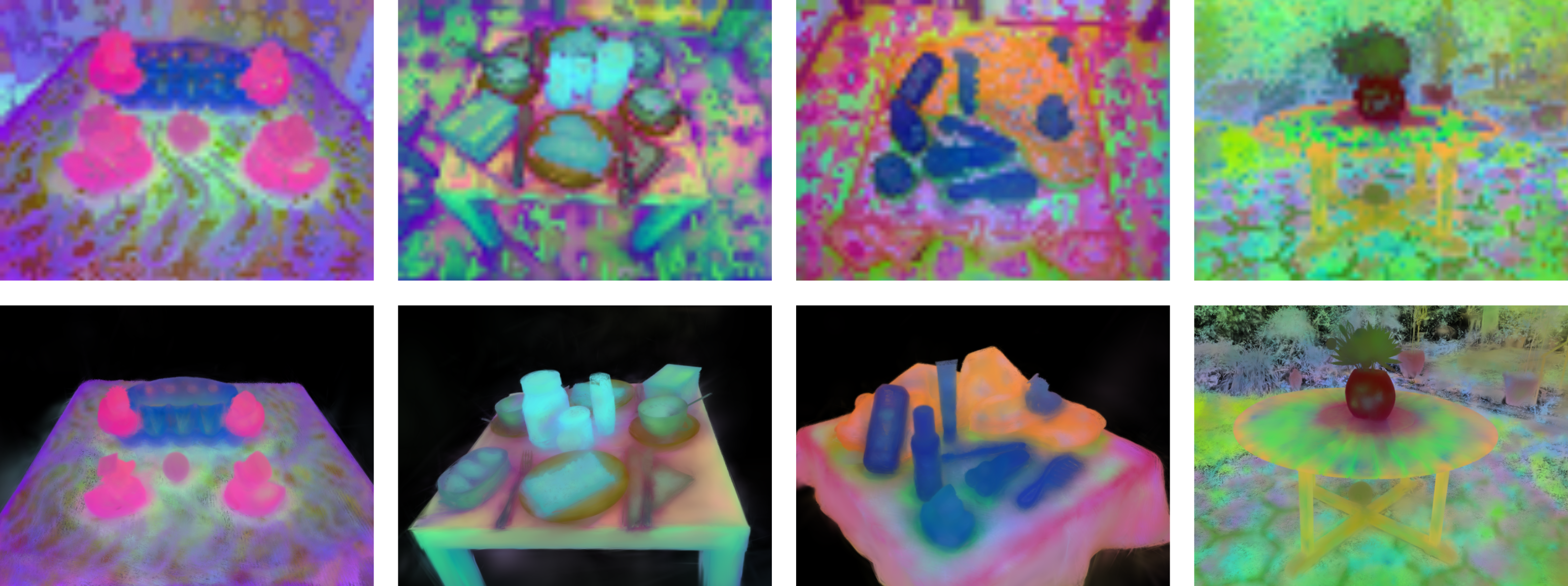}
   \caption{Scenes colorized with 3 PCA components of DINOv2 features. % to further motivate the applicability of our instant colorization approach in future downstream tasks. 
   Top: exemplary DINOv2 PCA images. Bottom: corresponding colorized 3DGS scenes with SH of order $L = 3$.}
   \label{fig:dino_dense}
\end{figure}

To demonstrate the potential of our instant colorization approach to accelerate these downstream applications, as a proof of concept, we apply it to quickly enrich a scene representation with neural image features. 
Specifically, DINOv2 features \cite{oquab2023dinov2} are extracted from the input images, principal components are computed, and the resulting feature maps are projected into the 3D Gaussian scene representation. 
\Cref{fig:dino_dense} shows exemplary DINOv2 PCA visualizations (top row) and the corresponding feature-enriched 3D scenes (bottom row) on our relighting dataset as well as the Mip-NeRF 360 garden scene (right).
We chose $L=3$ to capture the view dependence of neural features. Since the neural features are of fairly low resolution, no refinement steps were necessary and all of the shown feature mappings were computed in \emph{less than one second}.

% ich stecke das lieber mal in die conclusion / outlook...
% The efficiency of the proposed method results in low execution latencies, which are required, for example, in robotic tasks that involve frequent updates or re-projection of semantic feature maps.
% This includes open-vocabulary localization \cite{kruzhkov2026omcl}, articulated object tracking \cite{kerr2024rsrd}, language-guided navigation \cite{peiqi2024okrobot}, semantic exploration \cite{igelbrink2026lierex}, and higher-level reasoning or scene completion \cite{ha2022semabs}.
% The proposed projection method therefore enables the integration of semantically rich image features into efficient 3D scene representations.
% Alex: Eventuell Segmentation mit Feature Enrichment tauschen?
% Ich glaube von der Erklärreihenfolge wäre das besser
% Ich machs einfach mal. Wenns nicht passt könnt ihr es auch eifnach wieder zurück tauschen

\subsection{Segmentation}
\label{sec_segmentation}

We further evaluated the proposed colorization framework on a 3D scene segmentation task. % to demonstrate its applicability to semantic editing tasks. 
In this experiment, 2D segmentation masks were generated using the Segment Anything Model 2 (SAM2) \cite{ravi2025sam2} and projected back onto the 3D Gaussian splats to produce a consistent 3D segmentation. Since segmentation masks should not depend on the viewing direction, we set $L=0$ and no refinement steps were necessary. After the mask projection, all gaussians whose mask-values fell below a predefined threshold (0.6) were filtered out.

%\textcolor{red}{TODO: L=0, no refinement steps. masking of scene}

\paragraph{Setup}
%Segmentation masks were generated using the Segment Anything Model 2 (SAM2) \cite{ravi2025sam2,kirillov2023segany}.

\begin{figure}[h!]
    \centering
    \includegraphics[width=0.48\textwidth]{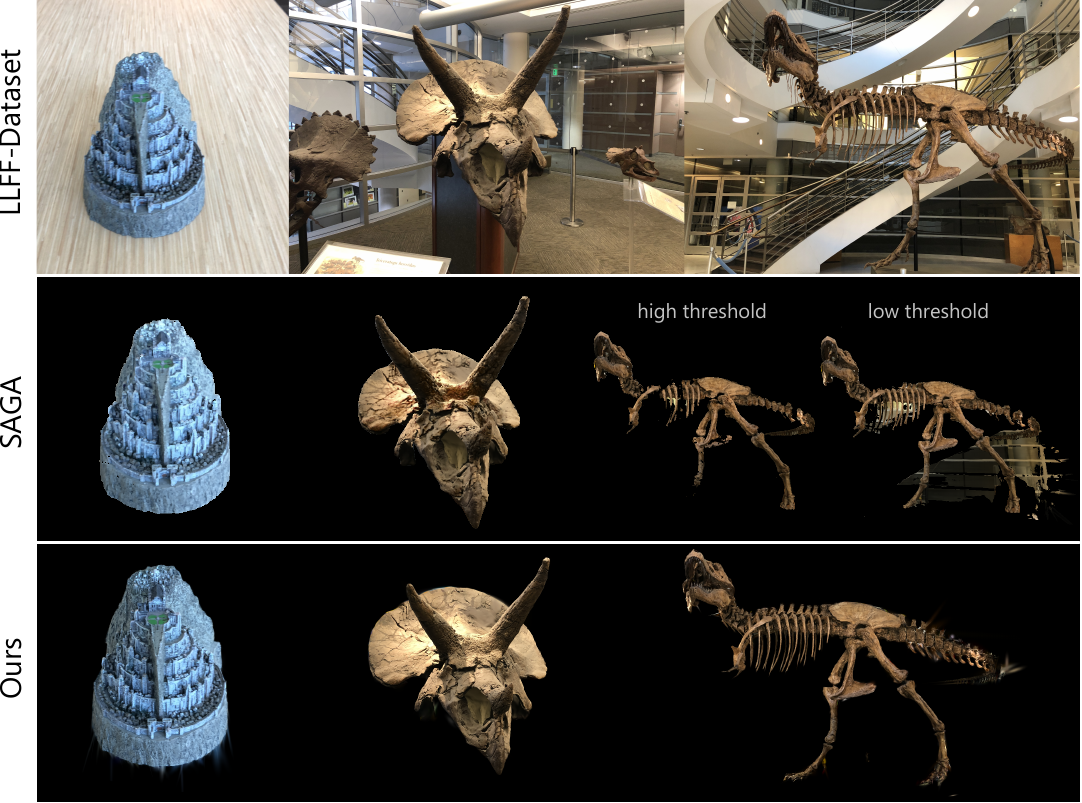}
    \caption{Segmentation results on LLFF dataset by SAGA \cite{cen2025saga} and our instant colorization approach. As can be seen by the T-Rex segmentation, our method excels at segmenting complex objects with fine structures.}
    \label{fig:segmentation_saga_ours}
\end{figure}

Although the inputs consisted of still images of a static scene, they were processed as a pseudo-video sequence to leverage SAM2’s temporal tracking and ensure consistent segmentation across all views. 
We used the \texttt{sam2\_hiera\_large.pt} checkpoint and its associated configuration \texttt{sam2\_hiera\_l.yaml}, running on a single NVIDIA 4090 GPU. 
%For initialization, points were placed manually on target objects in the first image (e.g., the table surface in \cref{fig:segment_table}). SAM2 then propagated this segmentation automatically through all remaining images, maintaining consistent object identifiers. 
%Each resulting frame was exported as a colorized overlay which was then used as the input for the colorization process (see \cref{fig:segmentation_qualitative}, left column and \cref{fig:segment_table}, top right).
For initialization, points were placed manually on target objects in the first image (e.g., the T-Rex skeleton in \Cref{fig:segmentation_saga_ours}). SAM2 then propagated this segmentation automatically through all remaining images. 
Each resulting frame was exported as a binary mask which was then used as the input for our instant colorization algorithm.

\paragraph{Results}

%As can be seen in Figures \ref{fig:segment_table}, bottom row and \ref{fig:segmentation_qualitative}, right column, projecting segmentation masks onto Gaussian splats can result in 3D segmentations that are spatially consistent across views and clearly delineate the target object.
We applied our segmentation approach to objects of various complexity in the LLFF dataset \cite{mildenhall2019llff} and compared the results to "Segment Any Gaussians" (SAGA) \cite{cen2025saga}, a state of the art segmentation approach for 3D gaussian splats.
As can be seen in \Cref{fig:segmentation_saga_ours}, for objects of low and medium complexity (see "Minas Tirith" on the left and "Triceratops skull" in the center), SAGA and our approach perform comparably. For objects of high complexity (see "T-Rex" on the right), however, our method achieves more accurate segmentation results for fine structures such as the ribs or the feet of the skeleton - even when finetuning SAGA's segmentation threshold and giving SAGA additional segmentation query points. 
Sometimes, our approach erroneously segments gaussians that penetrate the segmentation boundary. 
We believe this is mainly due to imperfections in the reconstruction and parts of the splats being invisible in the scene (see e.g. splats at the tail of the "T-Rex" or below "Minas Tirith" that would be usually covered by the table). 
While at inference time, SAGA is capable of producing segmentation masks in only 4ms, it requires an expensive contrastive learning step beforehand that takes about 20min on our machine. Our method on the other hand directly maps the segmentation masks of SAM2 onto the gaussian splats taking only around 1s. 
Additional segmentation results on our relighting dataset can be found in the supplementary.

\section{Conclusion}
\label{sec:conclusion}
In this work, we introduced a novel algorithm for efficient texture projection onto gaussian splats. The algorithm makes use of the normal equation to minimize a least squares problem % todo: etwas variabler schreiben
and can be easily implemented with existing differentiable GS renderers. 
Results on a new dataset of scenes under different lighting conditions and segmentation masks show that this algorithm can be effectively applied in various scene relighting, feature enrichment and semantic segmentation problems while significantly outperforming gradient descent based approaches such as Adam. 
%limitations + future work:
%- fast but not realtime % -> wo gäbe es denn noch potential für weitere beschleunigung?
% -> kleineres subset von bildern für projektion
% -> visibility check könnte genutzt werden um anzahl gaussians kleiner zu halten
The efficiency of the proposed method results in low execution latencies that could benefit numerous down-stream tasks, %, for example, in robotic tasks that involve frequent updates or re-projection of semantic feature maps.
%This has the potential to accelerate applications including 
such as open-vocabulary localization \cite{kruzhkov2026omcl}, articulated object tracking \cite{kerr2024rsrd}, language-guided navigation \cite{peiqi2024okrobot}, semantic exploration \cite{igelbrink2026lierex}, and higher-level reasoning or scene completion \cite{ha2022semabs}. 
%The proposed projection method therefore enables the integration of semantically rich image features into efficient 3D scene representations. 
Apart from scene relighting, feature enrichment and segmentation, we believe this algorithm might become useful for further texture projection tasks on gaussian splats such as scene stylization or texture painting. 
While our approach constitutes a significant acceleration for texture projections, it is not yet real-time capable. To gain further speed-ups, additional strategies could be employed such as reducing the number of projected camera images or clamping the number of Gaussians by considering their visibility to the observer.

{
    \small
    \bibliographystyle{ieeenat_fullname}
    \bibliography{main}

@String(CVPR= {IEEE Conf. Comput. Vis. Pattern Recog.})

@String(ICCV= {Int. Conf. Comput. Vis.})

@String(ECCV= {Eur. Conf. Comput. Vis.})

@String(NIPS= {Adv. Neural Inform. Process. Syst.})

@String(TOG= {ACM Trans. Graph.})

@String(ICLR = {Int. Conf. Learn. Represent.})

@String(AAAI = {AAAI})

@String(CVPR  = {CVPR})

@String(ICCV  = {ICCV})

@String(ECCV  = {ECCV})

@String(NIPS  = {NeurIPS})

@String(TOG   = {ACM TOG})

@String(ICLR  = {ICLR})

@inproceedings{mildenhall2020nerf,
 title={{NeRF}: Representing Scenes as Neural Radiance Fields for View Synthesis},
 author={Ben Mildenhall and Pratul P. Srinivasan and Matthew Tancik and Jonathan T. Barron and Ravi Ramamoorthi and Ren Ng},
 year={2020},
 booktitle=ECCV,
}

@article{mueller2022instant,
    author = {Thomas M\"uller and Alex Evans and Christoph Schied and Alexander Keller},
    title = {Instant Neural Graphics Primitives with a Multiresolution Hash Encoding},
    journal = {ACM Trans. Graph.},
    issue_date = {July 2022},
    volume = {41},
    number = {4},
    month = jul,
    year = {2022},
    pages = {102:1--102:15},
    articleno = {102},
    numpages = {15},
    url = {https://doi.org/10.1145/3528223.3530127},
    doi = {10.1145/3528223.3530127},
    publisher = {ACM},
    address = {New York, NY, USA}
}

@article{liu2019softras,
  title={Soft Rasterizer: A Differentiable Renderer for Image-based 3D Reasoning},
  author={Liu, Shichen and Li, Tianye and Chen, Weikai and Li, Hao},
  journal=ICCV,
  month = {Oct},
  year={2019}
}

@misc{laine2020modularprimitiveshighperformancedifferentiable,
  title={Modular Primitives for High-Performance Differentiable Rendering}, 
  author={Samuli Laine and Janne Hellsten and Tero Karras and Yeongho Seol and Jaakko Lehtinen and Timo Aila},
  year={2020},
  eprint={2011.03277},
  archivePrefix={arXiv},
  primaryClass={cs.GR},
  url={https://arxiv.org/abs/2011.03277}, 
}

@software{Mitsuba3,
    title = {Mitsuba 3 renderer},
    author = {Wenzel Jakob and Sébastien Speierer and Nicolas Roussel and Merlin Nimier-David and Delio Vicini and Tizian Zeltner and Baptiste Nicolet and Miguel Crespo and Vincent Leroy and Ziyi Zhang},
    note = {https://mitsuba-renderer.org},
    version = {3.1.1},
    year = 2022
}

@inproceedings{schoenberger2016sfm,
    author={Sch\"{o}nberger, Johannes Lutz and Frahm, Jan-Michael},
    title={Structure-from-Motion Revisited},
    booktitle=CVPR,
    year={2016},
}

@inproceedings{schoenberger2016mvs,
    author={Sch\"{o}nberger, Johannes Lutz and Zheng, Enliang and Pollefeys, Marc and Frahm, Jan-Michael},
    title={Pixelwise View Selection for Unstructured Multi-View Stereo},
    booktitle=ECCV,
    year={2016},
}

@article{tikhonov1977solutions,
  title={Solutions of ill posed problems},
  author={Tikhonov, Andrey Nikolayevich},
  year={1977},
  publisher={John Wiley \& Sons}
}

@inproceedings{Westover1991SplattingAP,
  title={Splatting: a parallel, feed-forward volume rendering algorithm},
  author={Lee Westover},
  year={1991},
  url={https://api.semanticscholar.org/CorpusID:18129593}
}

@inproceedings{ewa_volume_splatting,
author = {Zwicker, Matthias and Pfister, Hanspeter and van Baar, Jeroen and Gross, Markus},
title = {EWA volume splatting},
year = {2001},
isbn = {078037200X},
publisher = {IEEE Computer Society},
address = {USA},
booktitle = {Proceedings of the Conference on Visualization '01},
pages = {29–36},
numpages = {8},
location = {San Diego, California},
series = {VIS '01}
}

@INPROCEEDINGS{hw_ewa_splatting,
  author={Wei Chen and Liu Ren and Zwicker, M. and Pfister, H.},
  booktitle={IEEE Visualization 2004}, 
  title={Hardware-accelerated adaptive EWA volume splatting}, 
  year={2004},
  volume={},
  number={},
  pages={67-74},
  doi={10.1109/VISUAL.2004.38}}

@inproceedings{rhodin2015versatile,
    title={A Versatile Scene Model with Differentiable Visibility Applied to Generative Pose Estimation},
    author={Rhodin, Helge and Robertini, Nadia and Richardt, Christian and Seidel, Hans-Peter and Theobalt, Christian},
    booktitle = ICCV,
    month = {December},
    year = {2015}
}

@article{kerbl3Dgaussians,
      author       = {Kerbl, Bernhard and Kopanas, Georgios and Leimk{\"u}hler, Thomas and Drettakis, George},
      title        = {3D {G}aussian Splatting for Real-Time Radiance Field Rendering},
      journal      = {ACM Transactions on Graphics},
      number       = {4},
      volume       = {42},
      month        = {July},
      year         = {2023},
      url          = {https://repo-sam.inria.fr/fungraph/3d-gaussian-splatting/}
}

@inproceedings{Huang2DGS2024,
    title={2D {G}aussian Splatting for Geometrically Accurate Radiance Fields},
    author={Huang, Binbin and Yu, Zehao and Chen, Anpei and Geiger, Andreas and Gao, Shenghua},
    publisher = {Association for Computing Machinery},
    booktitle = {SIGGRAPH 2024 Conference Papers},
    year      = {2024},
    doi       = {10.1145/3641519.3657428}
}

@article{liu2025survey,
  title={A Survey of 3D Reconstruction: The Evolution from Multi-View Geometry to NeRF and 3DGS},
  author={Liu, Shuai and Yang, Mengmeng and Xing, Tingyan and Yang, Ran},
  journal={Sensors},
  volume={25},
  number={18},
  pages={5748},
  year={2025},
  publisher={MDPI}
}

@article{Dalal_2024,
   title={Gaussian Splatting: 3D Reconstruction and Novel View Synthesis: A Review},
   volume={12},
   ISSN={2169-3536},
   url={http://dx.doi.org/10.1109/ACCESS.2024.3408318},
   DOI={10.1109/access.2024.3408318},
   journal={IEEE Access},
   publisher={Institute of Electrical and Electronics Engineers (IEEE)},
   author={Dalal, Anurag and Hagen, Daniel and Robbersmyr, Kjell G. and Knausgård, Kristian Muri},
   year={2024},
   pages={96797–96820} 
}

@article{hahlbohm2025htgs,
  title   = {Efficient Perspective-Correct 3D {G}aussian Splatting Using Hybrid Transparency},
  author  = {Hahlbohm, Florian and Friederichs, Fabian and Weyrich, Tim and Franke, Linus and Kappel, Moritz and Castillo, Susana and Stamminger, Marc and Eisemann, Martin and Magnor, Marcus},
  journal = {Computer Graphics Forum},
  volume  = {44},
  number  = {2},
  year    = {2025},
  doi     = {10.1111/cgf.70014},
  url     = {https://fhahlbohm.github.io/htgs/}
}

@article{moenneloccoz2024,
author = {Moenne-Loccoz, Nicolas and Mirzaei, Ashkan and Perel, Or and de Lutio, Riccardo and Martinez Esturo, Janick and State, Gavriel and Fidler, Sanja and Sharp, Nicholas and Gojcic, Zan},
title = {3D Gaussian Ray Tracing: Fast Tracing of Particle Scenes},
year = {2024},
issue_date = {December 2024},
publisher = {Association for Computing Machinery},
address = {New York, NY, USA},
volume = {43},
number = {6},
issn = {0730-0301},
url = {https://doi.org/10.1145/3687934},
doi = {10.1145/3687934},
journal = {ACM Trans. Graph.},
month = nov,
articleno = {232},
numpages = {19}
}

@article{dontsplat,
    author = {Condor, Jorge and Speierer, Sebastien and Bode, Lukas and Bozic, Aljaz and Green, Simon and Didyk, Piotr and Jarabo, Adrian},
    title = {Don't Splat your {G}aussians: Volumetric Ray-Traced Primitives for Modeling and Rendering Scattering and Emissive Media},
    year = {2025},
    issue_date = {February 2025},
    publisher = {Association for Computing Machinery},
    address = {New York, NY, USA},
    volume = {44},
    number = {1},
    issn = {0730-0301},
    url = {https://doi.org/10.1145/3711853},
    doi = {10.1145/3711853},
    journal = {ACM Trans. Graph.},
    month = feb,
    articleno = {10},
    numpages = {17}
}

@InProceedings{hoellein_2025_3dgslm,
    title={{3DGS-LM}: Faster {G}aussian-Splatting Optimization with {L}evenberg-{M}arquardt},
    author={H{\"o}llein, Lukas and Bo\v{z}i\v{c}, Alja\v{z} and Zollh{\"o}fer, Michael and Nie{\ss}ner, Matthias},
    booktitle=ICCV,
    year={2025}
}

@INPROCEEDINGS{rump10groudtruth,
        author = {Rump, M. and Sarlette, R. and Klein, R.},
         title = {Groundtruth Data for Multispectral Bidirectional Texture Functions},
     booktitle = {CGIV 2010},
          year = {2010},
         month = {June},
         pages = {326--330},
      location = {Joensuu, Finland},
  organization = {Society for Imaging Science and Technology},
    bibdate = {2010-10-08},
  comment = {<a href="http://cg.cs.uni-bonn.de/en/publications/paper-details/rump-2010-spectralbtf/">project page</a>},
}

@article{mildenhall2019llff,
  title={Local Light Field Fusion: Practical View Synthesis with Prescriptive Sampling Guidelines},
  author={Ben Mildenhall and Pratul P. Srinivasan and Rodrigo Ortiz-Cayon and Nima Khademi Kalantari and Ravi Ramamoorthi and Ren Ng and Abhishek Kar},
  journal={ACM Transactions on Graphics (TOG)},
  year={2019}
}

@article{filip18evaluating,
title = {{Evaluating Physical and Rendered Material Appearance}},
author = {Filip, J. and Kolafov{\'{a}}, M. and Havl{\'\i}{\v c}ek, M. and
 V{\'a}vra, R. and Haindl, M. and Rushmeier H.},
journal = {The Visual Computer (Computer Graphics International 2018)},
issue = {},
number = {},
year = {2018},
publisher = {Springer},
DOI = {},
pages = {}
}

@article{Loshchilov_AdamW,
  author       = {Ilya Loshchilov and
                  Frank Hutter},
  title        = {Fixing Weight Decay Regularization in Adam},
  journal      = {CoRR},
  volume       = {abs/1711.05101},
  year         = {2017},
  url          = {http://arxiv.org/abs/1711.05101},
  eprinttype    = {arXiv},
  eprint       = {1711.05101},
  bibsource    = {dblp computer science bibliography, https://dblp.org}
}

@article{sun_light_stage_superres_2020,
author = {Sun, Tiancheng and Xu, Zexiang and Zhang, Xiuming and Fanello, Sean and Rhemann, Christoph and Debevec, Paul and Tsai, Yun-Ta and Barron, Jonathan T. and Ramamoorthi, Ravi},
title = {Light stage super-resolution: continuous high-frequency relighting},
year = {2020},
issue_date = {December 2020},
publisher = {Association for Computing Machinery},
address = {New York, NY, USA},
volume = {39},
number = {6},
issn = {0730-0301},
url = {https://doi.org/10.1145/3414685.3417821},
doi = {10.1145/3414685.3417821},
journal = {ACM Trans. Graph.},
month = nov,
articleno = {260},
numpages = {12}
}

@article{debevec2012light,
  title={The light stages and their applications to photoreal digital actors},
  author={Debevec, Paul},
  year={2012}
}

@inproceedings{chao2025texturedgaussians,
    title={Textured {G}aussians for Enhanced 3D Scene Appearance Modeling},
    author={Brian Chao and Hung-Yu Tseng and Lorenzo Porzi and Chen Gao and Tuotuo Li and Qinbo Li and Ayush Saraf and Jia-Bin Huang and Johannes Kopf and Gordon Wetzstein and Changil Kim},
    year={2025},
    booktitle=CVPR
}

@misc{huang2024texturedgsgaussiansplattingspatially,
    title={{Textured-GS}: {G}aussian Splatting with Spatially Defined Color and Opacity}, 
    author={Zhentao Huang and Minglun Gong},
    year={2024},
    eprint={2407.09733},
    archivePrefix={arXiv},
    primaryClass={cs.CV},
    url={https://arxiv.org/abs/2407.09733}, 
}

@article{chen2025quantifying,
  title={Quantifying and Alleviating Co-Adaptation in Sparse-View 3D {G}aussian Splatting},
  author={Chen, Kangjie and Zhong, Yingji and Li, Zhihao and Lin, Jiaqi and Chen, Youyu and Qin, Minghan and Wang, Haoqian},
  journal={arXiv preprint arXiv:2508.12720},
  year={2025}
}

@misc{chen2025snerfautonomousdrivingsimulation,
      title={S-NeRF++: Autonomous Driving Simulation via Neural Reconstruction and Generation}, 
      author={Yurui Chen and Junge Zhang and Ziyang Xie and Wenye Li and Feihu Zhang and Jiachen Lu and Li Zhang},
      year={2025},
      eprint={2402.02112},
      archivePrefix={arXiv},
      primaryClass={cs.CV},
      url={https://arxiv.org/abs/2402.02112}, 
}

@inproceedings{engelmann2024opennerf,
  title     = {{OpenNeRF: Open Set 3D Neural Scene Segmentation with Pixel-Wise Features and Rendered Novel Views}},
  author    = {Engelmann, Francis and Manhardt, Fabian and Niemeyer, Michael and Tateno, Keisuke and Pollefeys, Marc and Tombari, Federico},
  booktitle = {International Conference on Learning Representations},
  year      = {2024}
}

@article{Ming_2025,
   title={Benchmarking neural radiance fields for autonomous robots: An overview},
   volume={140},
   ISSN={0952-1976},
   url={http://dx.doi.org/10.1016/j.engappai.2024.109685},
   DOI={10.1016/j.engappai.2024.109685},
   journal={Engineering Applications of Artificial Intelligence},
   publisher={Elsevier BV},
   author={Ming, Yuhang and Yang, Xingrui and Wang, Weihan and Chen, Zheng and Feng, Jinglun and Xing, Yifan and Zhang, Guofeng},
   year={2025},
   month=jan, pages={109685} 
}

@misc{jurca2024rtgs2realtimegeneralizablesemantic,
      title={RT-GS2: Real-Time Generalizable Semantic Segmentation for 3D Gaussian Representations of Radiance Fields}, 
      author={Mihnea-Bogdan Jurca and Remco Royen and Ion Giosan and Adrian Munteanu},
      year={2024},
      eprint={2405.18033},
      archivePrefix={arXiv},
      primaryClass={cs.CV},
      url={https://arxiv.org/abs/2405.18033}, 
}

@article{yang2024spec,
  title={Spec-{G}aussian: Anisotropic View-Dependent Appearance for 3D {G}aussian Splatting},
  author={Yang, Ziyi and Gao, Xinyu and Sun, Yangtian and Huang, Yihua and Lyu, Xiaoyang and Zhou, Wen and Jiao, Shaohui and Qi, Xiaojuan and Jin, Xiaogang},
  journal=NIPS,
  year={2024}
}

@article{xu2024SuperGaussians,
    title={{SuperGaussians}: Enhancing {G}aussian Splatting Using Primitives with Spatially Varying Colors},
    author={Xu, Rui and Chen, Wenyue and Wang, Jiepeng and Liu, Yuan and Wang, Peng and Gao, Lin and Xin, Shiqing and Komura, Taku and Li, Xin and Wang, Wenping},
    year={2024},
    archivePrefix={arXiv},
    journal={arXiv},
    primaryClass={cs.LG},
}

@article{ye2025gsplat,
  title={gsplat: An open-source library for {G}aussian splatting},
  author={Ye, Vickie and Li, Ruilong and Kerr, Justin and Turkulainen, Matias and Yi, Brent and Pan, Zhuoyang and Seiskari, Otto and Ye, Jianbo and Hu, Jeffrey and Tancik, Matthew and Angjoo Kanazawa},
  journal={Journal of Machine Learning Research},
  volume={26},
  number={34},
  pages={1--17},
  year={2025}
}

@misc{kingma2017adammethodstochasticoptimization,
      title={Adam: A Method for Stochastic Optimization}, 
      author={Diederik P. Kingma and Jimmy Ba},
      year={2017},
      eprint={1412.6980},
      archivePrefix={arXiv},
      primaryClass={cs.LG},
      url={https://arxiv.org/abs/1412.6980}, 
}

@inproceedings{kirillov2023segany,
  title={Segment Anything},
  author={Kirillov, Alexander and Mintun, Eric and Ravi, Nikhila and Mao, Hanzi and Rolland, Chloe and Gustafson, Laura and Xiao, Tete and Whitehead, Spencer and Berg, Alexander C. and Lo, Wan-Yen and Doll{\'a}r, Piotr and Girshick, Ross},
  booktitle=ICCV,
  month= {October},
  year= {2023},
  pages= {4015-4026}
}

@inproceedings{
    ravi2025sam2,
    title={{SAM} 2: Segment Anything in Images and Videos},
    author={Nikhila Ravi and Valentin Gabeur and Yuan-Ting Hu and Ronghang Hu and Chaitanya Ryali and Tengyu Ma and Haitham Khedr and Roman R{\"a}dle and Chloe Rolland and Laura Gustafson and Eric Mintun and Junting Pan and Kalyan Vasudev Alwala and Nicolas Carion and Chao-Yuan Wu and Ross Girshick and Piotr Dollar and Christoph Feichtenhofer},
    booktitle=ICLR,
    year={2025},
    url={https://openreview.net/forum?id=Ha6RTeWMd0}
}

@article{oquab2023dinov2,
  title={Dinov2: Learning robust visual features without supervision},
  author={Oquab, Maxime and Darcet, Timoth{\'e}e and Moutakanni, Th{\'e}o and Vo, Huy and Szafraniec, Marc and Khalidov, Vasil and Fernandez, Pierre and Haziza, Daniel and Massa, Francisco and El-Nouby, Alaaeldin and others},
  journal={arXiv preprint arXiv:2304.07193},
  year={2023}
}

@article{cen2025saga, 
  title={Segment Any 3D Gaussians},
  volume={39},
  doi={10.1609/aaai.v39i2.32193},
  number={2},
  journal={Proceedings of the AAAI Conference on Artificial Intelligence},
  author={Cen, Jiazhong and Fang, Jiemin and Yang, Chen and Xie, Lingxi and Zhang, Xiaopeng and Shen, Wei and Tian, Qi},
  year={2025},
  month={Apr.},
  pages={1971-1979}
}

@inproceedings{debevec1998lightstage,
  author       = {Debevec, Paul E.},
  title        = {Rendering Synthetic Objects into Real Scenes: Bridging Traditional and Image-Based Graphics with Global Illumination and High Dynamic Range Photography},
  booktitle    = {Proceedings of ACM SIGGRAPH ’98},
  year         = {1998},
  pages        = {189–198},
  publisher    = {ACM},
  address      = {Orlando, FL, USA},
  doi          = {10.1145/280814.280882}
}

@misc{guo2024semanticgaussiansopenvocabularyscene,
    title={Semantic {G}aussians: Open-Vocabulary Scene Understanding with 3D {G}aussian Splatting}, 
    author={Jun Guo and Xiaojian Ma and Yue Fan and Huaping Liu and Qing Li},
    year={2024},
    eprint={2403.15624},
    archivePrefix={arXiv},
    primaryClass={cs.CV},
    url={https://arxiv.org/abs/2403.15624}, 
}

@inproceedings{Gao2024Relightable3DGaussian,
  author={Jian Gao and Chun Gu and Youtian Lin and Zhihao Li and Hao Zhu and Xun Cao and Li Zhang and Yao Yao},
  title={Relightable 3D {G}aussians: Realistic Point Cloud Relighting with {BRDF} Decomposition and Ray Tracing},
  year={2024},
  pages={73-89},
  booktitle=ECCV
}

@inproceedings{Kaleta2025LumiGauss,
  author = {Joanna Kaleta and Kacper Kania and Tomasz Trzciński and Marek Kowalski},
  title  = {{LumiGauss}: Relightable {G}aussian Splatting in the Wild},
  booktitle = {Proceedings of WACV 2025},
  year   = {2025}
}

@inproceedings{NormalGS2024,
  author = {Wei, Meng and Wu, Qianyi and Zheng, Jianmin and Rezatofighi, Hamid and Cai, Jianfei},
  title  = {3D {G}aussian Splatting with Normal-Involved Rendering},
  booktitle = NIPS,
  year   = {2024}
}

@inproceedings{Dihlmann2024SSSGS,
  author = {Jan-Niklas Dihlmann and Andreas Engelhardt and Arjun Majumdar and Raphael Braun and Hendrik P.A. Lensch},
  title  = {Subsurface Scattering for 3D {G}aussian Splatting},
  booktitle = NIPS,
  year   = {2024}
}

@article{Sun2025GRGS,
  author = {Yipengjing Sun and Chenyang Wang and Shunyuan Zheng and Zonglin Li and Shengping Zhang and Xiangyang Ji},
  title  = {Generalizable and Relightable {G}aussian Splatting for Human Novel View Synthesis},
  journal= {arXiv},
  year   = {2025},
  note   = {arXiv:2505.21502}
}

@inproceedings{garfield2024,
 author = {Kim, Chung Min and Wu, Mingxuan and Kerr, Justin and Tancik, Matthew and Goldberg, Ken and Kanazawa, Angjoo},
 title = {{GARField}: Group Anything with Radiance Fields},
 booktitle = {Conference on Computer Vision and Pattern Recognition (CVPR)},
 year = {2024},
}

@article{maggio2024clio,
  title={Clio: Real-time Task-Driven Open-Set 3D Scene Graphs},
  author={Maggio, Dominic and Chang, Yun and Hughes, Nathan and Trang, Matthew and
Griffith, Dan and Dougherty, Carlyn and Cristofalo, Eric and
Schmid, Lukas and Carlone, Luca},
  journal={IEEE Robotics and Automation Letters},
  year={2024},
  volume={9},
  number={10},
  pages={8921-8928},
  doi={10.1109/LRA.2024.3451395}
}

@article{peiqi2024okrobot,
  author={Peiqi, Liu and Yaswanth, Orru and Jay, Vakil and Chris, Paxton and Nur, Shafiullah and Lerrel, Pinto},
  title={Demonstrating OK-Robot: What Really Matters in Integrating Open-Knowledge Models for Robotics},
  journal={Robotics: Science and Systems},
  publisher={Robotics: Science and Systems Foundation},
  year={2024},
  month={07},
  doi={10.15607/RSS.2024.XX.091},
}

@article{conceptfusion2023,
  author    = {Jatavallabhula, {Krishna Murthy} and Kuwajerwala, Alihusein and Gu, Qiao and Omama, Mohd and Chen, Tao and Li, Shuang and Iyer, Ganesh and Saryazdi, Soroush and Keetha, Nikhil and Tewari, Ayush and Tenenbaum, {Joshua B.} and {de Melo}, {Celso Miguel} and Krishna, Madhava and Paull, Liam and Shkurti, Florian and Torralba, Antonio},
  title     = {{ConceptFusion}: Open-set Multimodal 3D Mapping},
  journal   = {Robotics: Science and Systems},
  year      = {2023},
}

@article{clipfields2023,
  author    = {Nur Muhammad Mahi Shafiullah and Chris Paxton and Lerrel Pinto and Soumith Chintala and Arthur Szlam},
  title     = {{CLIP}-Fields: Weakly Supervised Semantic Fields for Robotic Memory},
  journal   = {Robotics: Science and Systems},
  year      = {2023},
}

@inproceedings{peng2023openscene,
  title = {{OpenScene}: 3D Scene Understanding with Open Vocabularies},
  author = {Peng, Songyou and Genova, Kyle and Jiang, Chiyu "Max" and Tagliasacchi, Andrea and Pollefeys, Marc and Funkhouser, Thomas},
  booktitle = CVPR,
  year = {2023}
}

@inproceedings{ha2022semabs,
  title={Semantic Abstraction: Open-World 3{D} Scene Understanding from 2{D} Vision-Language Models},
  author = {Ha, Huy and Song, Shuran},
  booktitle={Proceedings of the 2022 Conference on Robot Learning},
  year={2022}
}

@inproceedings{lerf2023,
 author = {Kerr, Justin* and Kim, Chung Min* and Goldberg, Ken and Kanazawa, Angjoo and Tancik, Matthew},
 title = {{LERF}: Language Embedded Radiance Fields},
 booktitle = ICCV,
 year = {2023},
}

@inproceedings{kerr2024rsrd,
 title={Robot See Robot Do: Imitating Articulated Object Manipulation with Monocular 4D Reconstruction},
 author={Justin Kerr and Chung Min Kim and Mingxuan Wu and Brent Yi and Qianqian Wang and Ken Goldberg and Angjoo Kanazawa},
 booktitle={8th Annual Conference on Robot Learning},
 year={2024},
}

@article{kruzhkov2026omcl,
  author={Kruzhkov, Evgenii and Memmesheimer, Raphael and Behnke, Sven},
  journal={IEEE Robotics and Automation Letters}, 
  title={{OMCL}: Open-Vocabulary Monte Carlo Localization},
  year={2026},
  volume={},
  number={},
  pages={}
}

@misc{guenther2026sgb,
  title={A Scene Graph Backed Approach to Open Set Semantic Mapping}, 
  author={Martin Günther and Felix Igelbrink and Oscar Lima and Lennart Niecksch and Marian Renz and Martin Atzmueller},
  year={2026},
  eprint={2602.03781},
  archivePrefix={arXiv},
  primaryClass={cs.RO},
}

@article{igelbrink2026lierex,
  title={LIEREx: Language-Image Embeddings for Robotic Exploration},
  author={Igelbrink, Felix and Niecksch, Lennart and Renz, Marian and Günther, Martin and Atzmueller, Martin},
  year={2026},
  journal={KI - Künstliche Intelligenz},
  pages={1610-1987},
}

@misc{igelbrink2026disc,
      title={{DISC}: Dense Integrated Semantic Context for Large-Scale Open-Set Semantic Mapping}, 
      author={Felix Igelbrink and Lennart Niecksch and Martin Atzmueller and Joachim Hertzberg},
      year={2026},
      eprint={2603.03935},
      archivePrefix={arXiv},
      primaryClass={cs.CV},
      url={https://arxiv.org/abs/2603.03935}, 
}
}

% WARNING: do not forget to delete the supplementary pages from your submission 
% \input{sec/X_suppl}
%\input{sec/6_supplementary}

\end{document}